\documentclass[journal]{IEEEtran}

\usepackage{amsmath,amsfonts}
\usepackage{algorithm}
\usepackage{array}
\usepackage[caption=false,font=normalsize,labelfont=sf,textfont=sf]{subfig}
\usepackage{textcomp}
\usepackage{stfloats}
\usepackage{url}
\usepackage{verbatim}
\usepackage{graphicx}
\usepackage{cite}

\usepackage{booktabs}
\usepackage{orcidlink}
\usepackage{multirow}
\usepackage{pifont}
\newcommand{\xmark}{\ding{55}}%
\usepackage{graphicx}
\usepackage{algpseudocode}

\newcommand{\red}[1]{\textcolor{black}{#1}}
\usepackage{makecell}

\usepackage{multicol}
\newcommand{\customtitle}{
    \begin{center}
        {\LARGE \bfseries Supplementary Material for Masked Motion Diffusion Models} \\[1em]
        {\large Junkun~Jiang~\href{https://orcid.org/0000-0001-7478-001X}{\includegraphics[width=10pt]{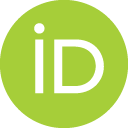}}, 
        Jie~Chen~\href{https://orcid.org/0000-0001-8419-4620}{\includegraphics[width=10pt]{figures/authors/orcid.png}},~\IEEEmembership{Senior Member,~IEEE}, 
        Ho~Yin~Au~\href{https://orcid.org/0000-0001-5588-4243}{\includegraphics[width=10pt]{figures/authors/orcid.png}}, 
        Jingyu~Xiang~\href{https://orcid.org/0009-0009-9035-4710}{\includegraphics[width=10pt]{figures/authors/orcid.png}}
} \\[1em]
    \end{center}
}

\begin{document}

\title{Learning Context-Adaptive Motion Priors \\for Masked Motion Diffusion Models \\with Efficient Kinematic Attention Aggregation}

\author{Junkun~Jiang~\href{https://orcid.org/0000-0001-7478-001X}{\includegraphics[width=10pt]{figures/authors/orcid.png}}, 
        Jie~Chen~\href{https://orcid.org/0000-0001-8419-4620}{\includegraphics[width=10pt]{figures/authors/orcid.png}},~\IEEEmembership{Senior Member,~IEEE}, 
        Ho~Yin~Au~\href{https://orcid.org/0000-0001-5588-4243}{\includegraphics[width=10pt]{figures/authors/orcid.png}}, 
        Jingyu~Xiang~\href{https://orcid.org/0009-0009-9035-4710}{\includegraphics[width=10pt]{figures/authors/orcid.png}}
\IEEEcompsocitemizethanks{\IEEEcompsocthanksitem J.~Jiang, J.~Chen (corresponding author), H.~Y.~Au and J.~Xiang are with the Department of Computer Science, Hong Kong Baptist University, Hong Kong SAR, China (e-mail: \{csjkjiang, chenjie, cshyau, csjyxiang\}@comp.hkbu.edu.hk).
\IEEEcompsocthanksitem This paper has supplementary
downloadable material available at http://ieeexplore.ieee.org, provided by the authors.
This includes a supplementary material with implementation details and more experimental results, and a demonstration video with visual comparisons to other state-of-the-art methods on motion quality.
This material is 51.9 MB in size.
}
}

\markboth{Manuscript accepted by IEEE Transactions on Multimedia}%
{Shell \MakeLowercase{\textit{et al.}}: A Sample Article Using IEEEtran.cls for IEEE Journals}

\maketitle

\begin{abstract}
Vision-based motion capture solutions often struggle with occlusions, which result in the loss of critical joint information and hinder accurate 3D motion reconstruction. Other wearable alternatives also suffer from noisy or unstable data, often requiring extensive manual cleaning and correction to achieve reliable results. To address these challenges, we introduce the Masked Motion Diffusion Model (MMDM), a diffusion-based generative reconstruction framework that enhances incomplete or low-confidence motion data using partially available high-quality reconstructions within a Masked Autoencoder architecture. Central to our design is the Kinematic Attention Aggregation (KAA) mechanism, which enables efficient, deep, and iterative encoding of both joint-level and pose-level features, capturing structural and temporal motion patterns essential for task-specific reconstruction. We focus on learning context-adaptive motion priors—specialized structural and temporal features extracted by the same reusable architecture, where each learned prior emphasizes different aspects of motion dynamics and is specifically efficient for its corresponding task. This enables the architecture to adaptively specialize without altering its structure. Such versatility allows MMDM to efficiently learn motion priors tailored to scenarios such as motion refinement, completion, and in-betweening. Extensive evaluations on public benchmarks demonstrate that MMDM achieves strong performance across diverse masking strategies and task settings. \red{The source code is available at \url{https://github.com/jjkislele/MMDM}.}
\end{abstract}


\section{Introduction}

\IEEEPARstart{M}{otion} capture technology (mocap) records high-fidelity human motion, which is widely applied in filmmaking, animation, and healthcare. Vision-based non-wearable mocap is attracting attention due to its affordability, user-friendliness, and wide accessibility. Utilizing 3D Human Pose Estimation (HPE) techniques, it can capture motion with standard RGB cameras, eliminating the need for specialized equipment.
Currently, HPE is mainly divided into two categories: \emph{1)} Multi-view Estimation~\cite{zhang20204d,reddy2021tessetrack,dong2021fast,jiang2022dual,choudhury2023tempo,jiang2025JCSAT,tu2020voxelpose,zhang2021lightweight,zhou2022quickpose}. This technique involves using multiple calibrated RGB cameras to simultaneously capture 2D image sequences of humans. It utilizes techniques such as 2D pose estimation, triangulation, re-identification, and tracking to achieve 3D motion reconstruction; \emph{2)} Monocular Estimation~\cite{pavllo20193d,zhang2022mixste,jiang2024exploring,zhang2023explicifying,xu2021graph}. This technique requires only a single RGB camera to map 2D pose estimation results into 3D space by establishing a re-projection relationship between 2D and 3D.
However, both of these encounter a fundamental challenge: insufficient motion capture accuracy, primarily due to occlusion issues. The missing information of key joints introduces ambiguity into data-driven human pose estimation models, resulting in deviations in predicting key joint positions and ultimately producing low-quality 3D reconstruction results. 
On the other hand, conditional motion generation has made significant progress in recent years. This technology can generate high-quality motions based on given conditions, such as music~\cite{au2022choreograph,au2024rechoreonet}, text~\cite{tevet2023mdm,karunratanakul2023gmd}, or other explicit signals~\cite{jiang2024motion,qi2024emotiongesture,au2025deep}. 
Therefore, it is feasible to integrate generation models into the 3D human motion estimation framework, using them to generate key joint positions that are unobservable due to occlusions. 

\begin{figure}[t]
\centering
\includegraphics[width=0.98\linewidth]{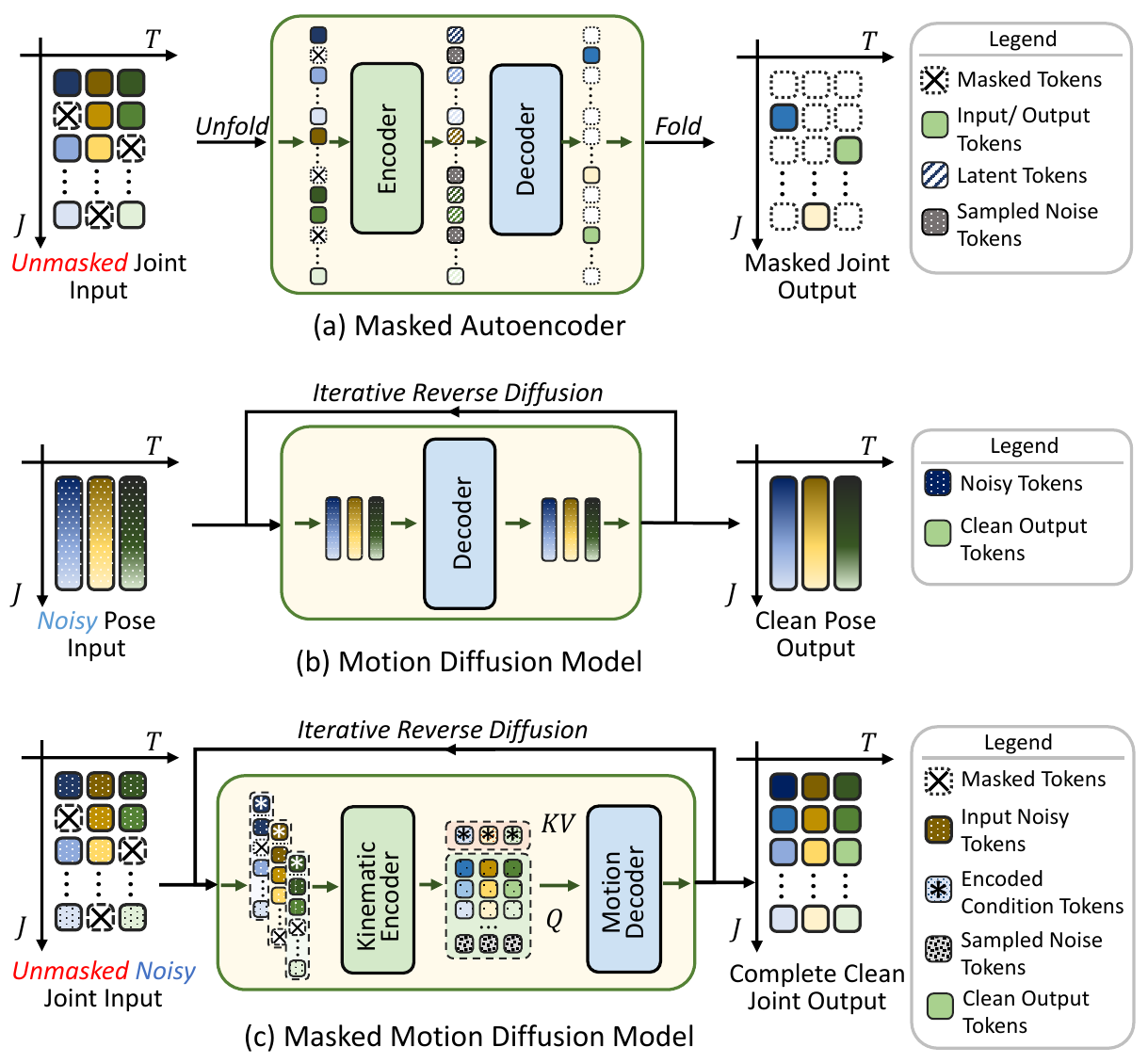}
\vspace{-10pt}
\caption{\red{Architecture comparison of the proposed Masked Motion Diffusion Model (MMDM) against other methods. \emph{(a)} Masked Autoencoders (MAEs)~\cite{jiang2022dual,yan2023skeletonmae,stoffl2024elucidating} reconstruct masked (low-confidence) joints from unmasked (visible/high-confidence) joints, but they are not designed for noisy input. \emph{(b)} Motion diffusion models~\cite{gong2023diffpose,kapon2024mas} denoise pose sequences to generate high-quality motions, which typically require complete input tokens.
\emph{(c)} Our MMDM combines both paradigms, taking partial, noisy inputs and fusing joint- and pose-level representations via the proposed Kinematic Attention Aggregation (KAA) to output complete, high-quality motions.}
}
\label{fig:model_diff}
\end{figure}

In the field of image reconstruction, Masked Autoencoder~\cite{he2022masked} has demonstrated its ability to globally reconstruct from partially masked pixel inputs, achieving breakthroughs in downstream tasks~\cite{he2022masked,wei2022masked,feichtenhofer2022masked,cheng2023forecast}. This approach shows that this self-supervised learning framework is able to provide a strong understanding of visual clues effectively. Recently, Masked Diffusion Model~\cite{wei2023diffusion} combines diffusion learning with the Masked Autoencoder structure, and experiments show that the reconstructed images have more details compared to the original MAE, highlighting the importance of generative models in reconstruction tasks. Inspired by these advancements, we extended our previous research D-MAE~\cite{jiang2022dual} by integrating the MAEs and diffusion models into motion modeling. As shown in Fig.~\ref{fig:model_diff}, traditional motion-based MAE can only reconstruct masked sets from unmasked ones, while the motion diffusion model processes noisy motion in its entirety. Our proposed Masked Motion Diffusion Model (MMDM) combines the advantages of both, enabling the generation of masked sets from unmasked noisy data. Therefore, by using well-reconstructed high-quality motions (unmasked sets) as conditions, we can generate the low-quality-filtered and missing parts (masked sets) through a conditional reverse diffusion process.

It is important to note that, unlike pixel-level representation in images, human motion involves two distinct dimensions of signals: spatial (body skeletal structure) and temporal (key joints' trajectory). Therefore, for motion modeling, as shown in Table~\ref{tab:motion_representation}, HPE-related works~\cite{jiang2022dual,gong2023diffpose,jiang2025JCSAT,li2022exploiting} tend to extract motion representations using key joints from both spatial and temporal dimensions, which we refer to as joint-level representation. Normally, they employ two transformer-based encoders to sequentially encode motion in the skeleton and trajectory dimensions. In contrast, for motion generation, particularly diffusion learning-based methods~\cite{tevet2023mdm,karunratanakul2023gmd}, extracting motion representation at the pose level is preferred. A single encoder is used to encode motion. 
\red{Although joint-level representations enable independent modeling of individual joints and capture rich spatial-temporal correlations, their application in diffusion models leads to substantially increased computational costs.
To address this challenge, we propose the Kinematic Attention Aggregation (KAA) mechanism, which integrates joint-level and pose-level features and models their spatio-temporal dependencies adaptively and efficiently.}

Overall, this innovative framework introduces the integration of a generative model within the reconstruction paradigm to enhance motion capture data. Leveraging the efficient KAA representation fusion and the inherent flexibility of MAEs, our model is adaptable to a variety of tasks.
\red{To the best of our knowledge, our MMDM represents the pioneer attempt to jointly fuse joint-level and pose-level representations while incorporating a generative reconstruction framework into the motion capture domain.} The main contributions of our work are summarized as follows: 
\begin{itemize}
\item We propose the Kinematic Attention Aggregation mechanism (KAA), which efficiently combines joint-level and pose-level information to enable iterative and deep encoding of spatio-temporal motion features. This approach enhances the representation of fine-grained dynamics and global coherence, while maintaining computational efficiency.
\item By integrating KAA with a masked diffusion paradigm, we introduce the Masked Motion Diffusion Model (MMDM)—a generative reconstruction framework that utilizes partial, high-quality motion data for conditional generation of incomplete motion capture data, demonstrating strong contextual understanding of incomplete motion sequences.
\item We show that the proposed KAA-based architecture can learn context-adaptive motion priors tailored to different tasks, efficiently capturing diverse structural and temporal aspects of motion dynamics without requiring architectural changes. This versatility is validated across motion completion, refinement, and in-betweening tasks, where our method consistently achieves strong results on public motion capture benchmarks.
\end{itemize}

\setlength{\tabcolsep}{2.0pt}
\begin{table}[!t]
\begin{center}
\caption{Comparisons of motion representation in state-of-the-art studies, with their objectives: human pose estimation (HPE), motion in-betweening (MIB), and motion generation (MG).}
\vspace{-7pt}
\begin{tabular}{rcccccc}
\toprule
\multirow{2}{*}{Method} & \multirow{2}{*}{Pub. Venue} & 
\multirow{2}{*}{Joint lvl.} & \multirow{2}{*}{Pose lvl.} & \multicolumn{3}{c}{Objective} \\
\cmidrule(lr){5-7}
& & & & HPE & MIB & MG \\
\midrule
D-MAE~\cite{jiang2022dual} & ACMMM'22 & $\checkmark$ & & $\checkmark$ & & \\
StridedTran.~\cite{li2022exploiting} & TMM'22 & $\checkmark$ & & $\checkmark$ & & \\
DiffPose~\cite{gong2023diffpose} & CVPR'23 & $\checkmark$ & & $\checkmark$ & & \\
MDM~\cite{tevet2023mdm} & ICLR'23 & & $\checkmark$ & & $\checkmark$ & $\checkmark$ \\
GMD~\cite{karunratanakul2023gmd} & ICCV'23 & & $\checkmark$ & & $\checkmark$ & $\checkmark$ \\
JCSAT~\cite{jiang2025JCSAT} & TVCG'25 & $\checkmark$ & & $\checkmark$ & & \\
\midrule
MMDM (Ours) & - & $\checkmark$ & $\checkmark$ & $\checkmark$ & $\checkmark$ & $\checkmark$ \\
\bottomrule
\vspace{-8mm}
\end{tabular}
\end{center}
\label{tab:motion_representation}
\end{table}

\section{Related work}

\noindent\textbf{3D Human Pose Estimation.} 
Existing 3D Human Pose Estimation (HPE) can be roughly categorized into two groups: multi-view 3D reconstruction and monocular 2D-to-3D lifting. The former approaches~\cite{belagiannis20153d,zhang20204d,zhang2021lightweight,lin2021multi,zhou2022quickpose,srivastav2024selfpose3d,reddy2021tessetrack,tu2020voxelpose} mainly rely on triangulation and focus on cross-view association. 
4D Association Graph (4DAG)~\cite{zhang20204d} proposes an energy function for associating 2D key joints, optimizing their connectivity, while MVPose~\cite{zhang2021lightweight} associates them using re-identification features to improve identity consistency across the views. JCSAT~\cite{jiang2025JCSAT} suggests ignoring the identities of 2D detections and triangulating them as unlabeled 3D point clouds, and utilizing a Transformer to explore the motion kinematics to get the 3D motion behind them. Some approaches~\cite{reddy2021tessetrack,choudhury2023tempo} project 2D heatmaps into a 3D volume as 3D heatmaps and directly regress them to 3D key joint positions. 
On the other side, monocular lifting methodologies~\cite{xu2021graph,zhang2022mixste,gong2023diffpose,pavllo20193d,li2022exploiting,rempe2021humor} concentrate on learning the implicit projection from 2D to 3D. 
Some methods~\cite{zhao2022graformer,xu2021graph} integrate Graph Convolutional Networks (GCNs) to explore the kinematic structure of humanoid movements. Some methods~\cite {zhao2022graformer,zhang2022mixste} incorporate attention mechanisms~\cite{vaswani2017attention} into GCNs, enabling them to capture more implicit relations than a single adjacency matrix can represent.  

\noindent\textbf{Masked Autoencoder.}  Inspired by the foundational study by He et al.~\cite{he2022masked}, researchers apply this unique paradigm for various vision-based reconstruction tasks, such as image feature completion~\cite {wei2022masked}, video completion~\cite{feichtenhofer2022masked,cheng2023forecast}, and point cloud completion~\cite{jiang2023masked,liu2023inter}.
Analogous to the relationship between images and videos, motion can be seen as another temporal signal with a stronger structural prior. 
To effectively model it, SkeletonMAE~\cite{yan2023skeletonmae} employs a GCN backend for 2D skeleton reconstruction, which is further applied to 2D motion recognition, resulting in significant improvements.
Mascaro et al.~\cite{mascaro2023unified} investigate MAEs on motion completion and denoising tasks.
Our earlier research, D-MAE~\cite{jiang2022dual}, incorporates MAEs into motion capture systems to fill in missing motion data. 
However, this study does not extend to evaluating its effectiveness for other downstream tasks. 

\noindent\textbf{Diffusion-based Motion Modeling.} 
Some studies~\cite{choi2022diffupose,gong2023diffpose} consider the depth ambiguity as noise and utilize diffusion models to improve the 2D-to-3D lifting results. Gong et al.~\cite{gong2023diffpose} employ a GCN structure and accelerate the diffusion process using DDIM~\cite{song2021denoising}. They assume that the noisy 3D input is sampled from an indeterminate pose distribution. We follow this assumption and utilize the diffusion model to refine motion capture data iteratively. MMM~\cite{pinyoanuntapong2024mmm} utilizes masked learning with diffusion models for text-based motion generation by masking low-confidence pose tokens. 
\red{Chen~\cite{chen2024text} further explores this direction by introducing frame-wise and body-part masking strategies to improve contextual reasoning among spatio-temporal semantics.
In contrast to these approaches, our model operates at the joint level and aggregates information to the pose level using the KAA mechanism. This design enables more effective modeling of kinematic dependencies.}

\noindent\textbf{Motion In-betweening.} Generally, it can be categorized into two groups. The first group~\cite{harvey2018recurrent,harvey2020RMIB} relies on autoregressive frame generation, producing transitioning frames to connect the boundaries of preceding and succeeding segments. The second group~\cite{kim2022CMIB,tevet2023mdm,karunratanakul2023gmd} relies on variable-length transitional segment generation. They incorporate text-based conditions to ensure that the context of the transitioning segment aligns with both the preceding and succeeding segments.

\begin{figure*}
\centering
\includegraphics[width=0.95\linewidth]{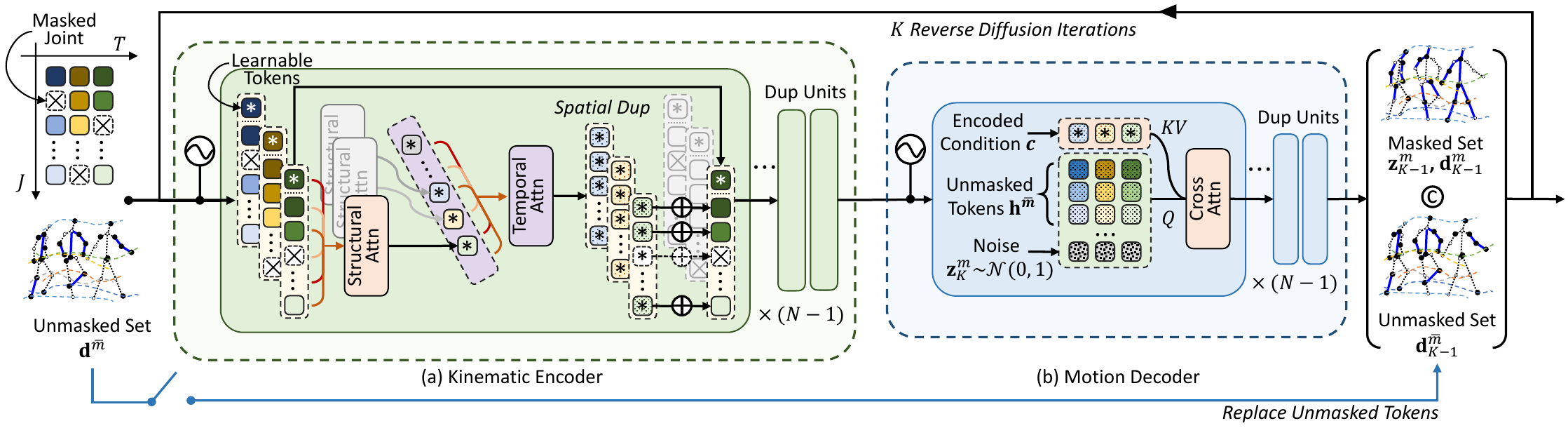}
\vspace{-10pt}
\caption{
Illustration of the reverse diffusion process in the proposed Masked Motion Diffusion Model (MMDM). It begins at iteration $k=K$ and proceeds sequentially to $k=0$, reconstructing the masked motion sequence $\mathbf{d}^{m}_k$ from Gaussian noise into high‐quality data, conditioned on the unmasked motion sequence $\mathbf{d}^{\overline{m}}$. Specifically, at each iteration $k$, \emph{(a)} the Kinematic Encoder encodes the unmasked joints to latent tokens $\mathbf{h}^{\overline{m}}$ and yields the kinematic condition $c$, and \emph{(b)} the Motion Decoder decodes the concatenated tokens $[\mathbf{h}^{\overline{m}}; \mathbf{z}_{k}^{m}]$ conditioned on $c$. To preserve motion context, the unmasked output $\mathbf{d}^{\overline{m}}_{k}$ is replaced by the input $\mathbf{d}^{\overline{m}}$ at every step. Positional encoding involves joint indices, frame numbers, and diffusion step indices are incorporated into the hidden features before each encoding and decoding stage.
}
\vspace{-10pt}
\label{fig:system}
\end{figure*}

\section{Methodology}

In this section, we begin by examining the fundamentals of the Masked Autoencoder, the Diffusion Model, and the proposed Masked Autoencoder-based Diffusion Model (MMDM) in Sec.~\ref{sec:preliminary}. 
Next, we elaborate on the architecture design of our MMDM in Sec.~\ref{sec:mdl}, where the diffusion learning strategy is integrated into the MAE decoding paradigm, allowing the model to approximate the motion distribution effectively. 
\red{Within this framework, we highlight the Kinematic Attention Aggregation (KAA) mechanism—a key component that efficiently fuses joint-level and pose-level motion representations, enhancing both fine-grained dynamic features and global motion coherence. Owing to KAA, our MMDM demonstrates robust motion modeling capabilities while maintaining computational efficiency, and supports multiple motion-related tasks:}
\textit{(1) Motion Completion}, which reconstructs masked joint sequences, in Sec.~\ref{sec:mot_compl}. 
\textit{(2) Motion Refinement}, which refines the existing motion data through the reverse diffusion process, in Sec.~\ref{sec:mot_refine}; 
\textit{(3) Motion In-betweening}, which treats the transitioning motion as a masked segment, using preceding and succeeding motions to reconstruct it, in Sec.~\ref{sec:mib}.

\subsection{Preliminary}\label{sec:preliminary}

\paragraph{Motion-based Masked Autoencoder}
Given a motion sequence $\mathbf{d} \in \mathbb{R}^{T \times J \times d}$ comprising $T$ frames with $J$ key joints per frame, each represented in $d$ dimensions, the objective is to reconstruct the masked motion data using the unmasked motion data.
Following D-MAE~\cite{jiang2022dual}, the process can be formulated as:
\begin{align}
\mathbf{h}^{\overline{m}} &= E_\phi(\mathbf{d}^{\overline{m}}) , \\
\widehat{\mathbf{d}} &= D_\theta( \mathbf{h}^{\overline{m}} +\!\!\!+ \: \mathbf{z}^m ) , \label{eq:mae} \\
\ell_{rec} &= \left\| {\widehat{\mathbf{d}}}^m - \mathbf{d}^m \right\|_2,
\end{align}
where $\left ( \cdot \right )^m$ refers to the masked set, while $\left ( \cdot \right )^{\overline{m}}$ denotes the unmasked set. The masking operation is applied across both the $T$ and $J$ dimensions. The encoder $E_{\phi}$ and the decoder $D_{\theta}$ are parameterized by $\phi$ and $\theta$, respectively. $E_{\phi}$ extracts hidden features $\textbf{h}^{\overline{m}}$ from the unmasked set, while $\mathbf{z}^m$ represents learnable embeddings for the masked set, initialized from a normal distribution. These embeddings $\mathbf{z}^m$ are concatenated with $\textbf{h}^{\overline{m}}$ using the $ +\!\!+ $ operator, and $D_{\theta}$ subsequently reconstructs the motion sequence $\widehat{\mathbf{d}}$. 

The reconstruction objective $\ell_{rec}$ employs the L2-norm to compute loss only over the masked joints, enforcing $E_{\phi}$ to infer rich semantic information from incomplete motion sequences. This empowers $D_{\theta}$ to faithfully reconstruct the missing data. As a result, the encoder’s learned representations acquire strong prior properties, effectively supporting a variety of downstream tasks, such as motion completion~\cite{jiang2022dual} and action recognition~\cite{yan2023skeletonmae}. 

\paragraph{Diffusive Probabilistic Modeling}
Diffusion models~\cite{ho2020denoising,song2021denoising} consider the transition between the input image $\mathbf{x}_0$ and the Gaussian noise $\mathbf{x}_K \sim \mathcal{N}(0, \mathbf{I})$ as a Markov chain which can be learned by a probabilistic generative model $G_\varphi$, parameterized by $\varphi$, under a fixed schedule configured by the diffusion step $K$. The chain-like process $\left ( \textbf{x}_0 \rightarrow \textbf{x}_1 \rightarrow \cdots \rightarrow \mathbf{x}_K \right )$ is called \textit{forward diffusion} and the converse process is called \textit{reverse diffusion}. Specifically, given an input image $\textbf{x}_0$ sampled from the training distribution $\textbf{x}_0 \sim q(\textbf{x})$, as suggested by \cite{ho2020denoising}, \textit{forward diffusion} can be formulated as: 
\begin{align}
\mathbf{x}_k &= \sqrt{\bar{\alpha}_k}\mathbf{x}_0 + \sqrt{1 - \bar{\alpha}_k}\boldsymbol{\epsilon} , \label{eq:alpha_blend} \\
q(\mathbf{x}_k \vert \mathbf{x}_0) &= \mathcal{N}(\mathbf{x}_k; \sqrt{\bar{\alpha}_k} \mathbf{x}_0, (1 - \bar{\alpha}_k)\mathbf{I}) ,
\end{align}
where $\epsilon \sim \mathcal{N}(\mathbf{0}, \mathbf{I})$, $\alpha_t = 1 - \beta_k$ and $\bar{\alpha}_k = \prod_{i=1}^k \alpha_i$, with a variance schedule $ \{\beta_k \in (0, 1)\}_{k=1}^K $. When $\beta_{K} \to 1$, leading to $\bar{\alpha}_K \to 0$, the last step of the \textit{forward diffusion} which approximates Gaussian noise $\epsilon$. Conversely, \textit{reverse diffusion} follows $\left ( \textbf{x}_0 \leftarrow \textbf{x}_1 \leftarrow \cdots \leftarrow \mathbf{x}_K \right )$ by an alpha blending between $\mathbf{x}_0$ and $\epsilon$ with the scaling schedule $\beta_k$.
During inference, the model recurrently performs $\mathbf{x}_{k-1}=G_\varphi(\mathbf{x}_{k}, k)$ and finally generates a high-quality sample $\mathbf{x}_0$.

\paragraph{Diffusion Modeling for Masked Motion Completion}
Building upon the above assumptions and definitions, we aim to model the distribution of the masked motion $\mathbf{d}^m$ conditioned by the unmasked motion $\mathbf{d}^{\overline{m}}$. 
Through reverse diffusion, $\mathbf{d}^m_0$ is progressively denoised, starting from an initial state $\mathbf{d}^m_{K} \sim \mathcal{N}(\mathbf{0}, \mathbf{I})$.  
At each iteration $k$, the encoder $E_{\phi}$ extracts the kinematic condition $c$ (see Sec.~\ref{sec:mdl}) from $\mathbf{d}^{\overline{m}}_{k}$, and the decoder $D_{\theta}$ generates a less noisy state $\mathbf{d}^m_{k-1}$. To preserve the overall motion context, the unmasked motion is consistently restored to its original values at each step.
In summary, the procedure integrates Eq.~\ref{eq:mae} and Eq.~\ref{eq:alpha_blend}, with $\mathbf{d}^m_k$ substituted for $\mathbf{x}_k$ and $D_{\theta}$ replacing $G_{\varphi}$, as formulated below:
\begin{align}
\mathbf{h}^{\overline{m}},c &= E_\phi(\mathbf{d}^{\overline{m}}),
\\
\widehat{\mathbf{d}}_{k-1}^{\overline{m}}, \widehat{\mathbf{d}}_{k-1}^m &= D_\theta(\mathbf{h}^{\overline{m}} +\!\!\!+ \: \mathbf{z}_{k}^m, c, k).
\end{align}
Following the standard objective formulation~\cite{rombach2021highresolution,wei2023diffusion}, we optimize the model using:
\begin{equation}
{\ell}_k = \mathbb{E}_{k \sim \{1, K \}, \mathbf{d}^m_k} \big\Vert \mathbf{d}^m_k - \widehat{\mathbf{d}}^m_k \big\Vert_2.
\end{equation} 

\subsection{Masked Motion Diffusion Model}\label{sec:mdl}
The proposed MMDM is built on an autoencoder structure, featuring a Kinematic Encoder with a \red{Kinematic Attention Aggregation (KAA) mechanism}, and a conventional Transformer-based Motion Decoder. As illustrated in Fig.~\ref{fig:system}, the Kinematic Encoder consists of $N$ pairs of self-attention blocks—each pair containing one Structural Attention block and one Temporal Attention block. \red{These blocks are interconnected by the KAA mechanism.} The Motion Decoder, in turn, is composed of $N$ cross-attention blocks.

During reverse diffusion, starting at step $k=K$, the unmasked input $\mathbf{d}^{\overline{m}}$ is projected and \red{processed by the Kinematic Encoder to produce the unmasked motion tokens $\mathbf{h}^{\overline{m}}$ and the kinematic condition $c$ via $N$ rounds of KAA process.
Within the encoder, data is initially processed by the Structural Attention block along the joint dimension, which is designed to focus on joint-level modeling and to extract structural features. These features are aggregated into pose-level representations and subsequently processed by the Temporal Attention block along the temporal dimension to capture trajectory information. The KAA mechanism effectively bridges the two blocks, enabling comprehensive information exchange across both structural and temporal axes, and facilitating the integration of joint- and pose-level motion representations.
}

\paragraph{Kinematic Attention Aggregation}
To obtain a representative and efficient embedding across the skeletal structure and the kinematics, \red{we introduce the Kinematic Attention Aggregation (KAA) mechanism. This mechanism operates within the Kinematic Encoder, interconnecting each pair of Structural Attention and Temporal Attention blocks to facilitate the fusion of joint- and pose-level representations. Below, we detail the processing steps of KAA.}

Suppose there is a motion latent embedding $\textbf{h} = \{h_{t} \in \mathbb{R}^{J\times D} \}_{t=0}^{T-1}$ consists of key frame pose $h_{t}$ with $D$ hidden dimension (for clarity, the unmask indicator is omitted here). As shown in Fig.~\ref{fig:system}(a),  $\textbf{h}$ (depicted as squares), is attached with an additional set of learnable embeddings, $\textbf{h}^{*} = \{h^{*}_{t} \in \mathbb{R}^{1\times D} \}_{t=0}^{T-1}$ (shown as star-marked squares).
This sequence $\{[h_t^*; h_t] \in \mathbb{R}^{(1+J)\times D}\}_{t=0}^{T-1}$ is processed by a Structural Attention block along the joint ($J$) dimension, enabling aggregation of structural information from each pose $h_{t}$ into the corresponding $h^{*}_{t}$. The learnable vector $h^{*}_{t}$ is designed to emulate pose-level representations, consistent with previous motion inbetweening/generation frameworks~\cite{tevet2023mdm,karunratanakul2023gmd}.

\red{Subsequently, \textbf{only} $\textbf{h}^{*}$ is processed through a Temporal Attention block along the temporal ($T$) dimension, thereby exploring trajectory-level dependencies.
It is important to emphasize that, in contrast to previous HPE studies~\cite{jiang2022dual,li2022exploiting,zhang2022mixste} which utilize self-attention over the $J$ dimension, our method substantially reduces computational complexity while still ensuring thorough spatial and temporal information exchange. This design enhances the model’s ability to capture relevant correlations efficiently.}

\red{Finally, after the process of Structure Attention and Temporal Attention blocks, the learnable embeddings $\textbf{h}^{*}$ are duplicated along the joint dimension and added back to the original latent embeddings $\textbf{h}$, marking the completion of one round of the KAA process. After $N$ rounds, the final $\textbf{h}^{*}$ is adopted as the kinematic condition $c$, encapsulating the full motion context needed for effective conditional decoding.}

\paragraph{Positional Encoding}
Following D-MAE~\cite{jiang2022dual}, we apply the Fourier positional embeddings~\cite{tancik2020fourier} to preserve skeletal and temporal information as well as the spatial relationship among key joints. For diffusion time step integration, we adopt the standard sinusoidal encoding method~\cite{song2021denoising}. These positional encodings are added to hidden features before each processing step in the encoder and decoder.

\begin{figure*}[!t]
\centering
\includegraphics[width=0.98\linewidth]{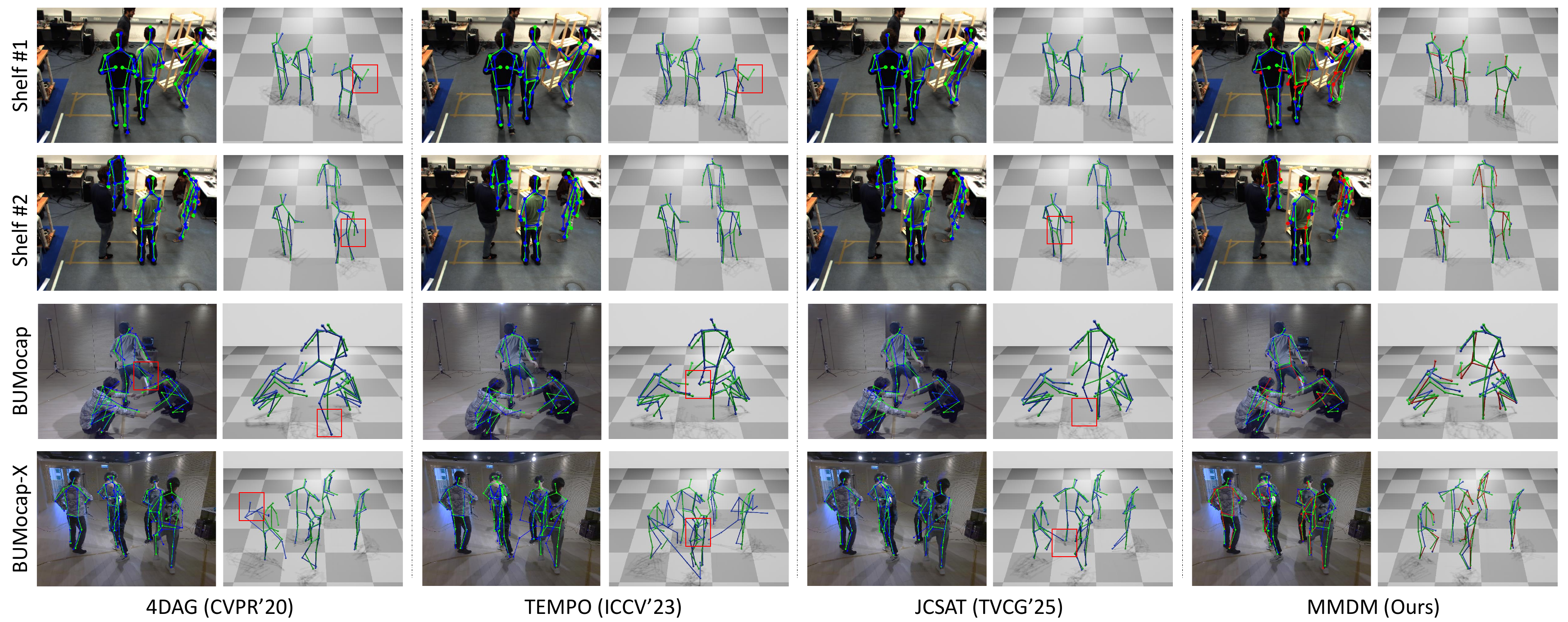}
\vspace{-10pt}
\caption{Qualitative comparisons of motion capture performance. Both 2D projections and new perspectives of 3D renderings are shown. In other methods, green and blue skeletons represent the ground truth and motion capture results, respectively. For our approach, applied in the motion completion setting, green, blue, and red skeletons differentiate the ground truth, motion capture results, and completion results. Red dotted boxes highlight the failure cases, indicating that our method yields more accurate results. }
\label{fig:result_shelf}
\end{figure*}

\section{Applications}

\subsection{Motion Completion}\label{sec:mot_compl}

Completion of missing data is the most straightforward application of MAEs. To validate the MMDM, we integrate it into a motion capture system to enhance performance by completing masked motion data. Initially, a baseline motion capture system is established to produce low-quality motion data. Next, a masking algorithm is developed to identify key joints needing masking due to low quality or being missing. Finally, the MMDM completes the missing or masked data, resulting in high-quality motion data.

\paragraph{A Simple yet Robust Mocap Framework}
Following a common implementation~\cite{dong2021fast,lin2021multi,jiang2022dual,jiang2025JCSAT}, we adopt the two-stage pipeline to reconstruct motion sequences. 
Suppose there are $N$ people recorded by $V$ calibrated RGB cameras within $T$ frames. The estimated 2D key joint locations are notated as $\mathbf{p} \in \mathbb{R}^{N \times V \times T \times J \times 2}$ with corresponding confidence scores $\rho \in [0, 1]^{N \times V \times T \times J \times 1} $, where $\rho=0$ indicates invisibility. 
To achieve geometric consistency across views, identity matching is performed using mid-hip epipolar distances and optimized via the Hungarian algorithm. The matched 2D poses are then triangulated to yield 3D motions $\mathbf{d} \in \mathbb{R}^{N \times T \times J \times 3}$ with triangulation errors $\sigma \in [0, 1]^{N \times T \times J \times 1} $. Identity tracking is incorporated to ensure consistency throughout the sequence.

\paragraph{Adaptive Masking Process}
We adaptively mask 3D key joints when they are occluded, have low 2D estimation confidence, or have high triangulation error. The masking probability for each joint is determined by a weight $w_j^t$ based on 2D confidence score $\rho_{j,v}^{t}$ and triangulation error $\sigma_{j}^{t}$:
\begin{equation}\label{eq:fuse}
w_{j,t} = \omega \cdot e^{-\sum_{v=1}^{V}\rho_{j, v}^{t}} + \sigma_{j}^{t}.
\end{equation} 
Here, the coefficient $\omega$ balances the influence of confidence and error. This equation ensures joints with lower confidence or higher triangulation error receive larger weights, thereby increasing their likelihood of being masked.

\subsection{Motion Refinement}\label{sec:mot_refine}
This task aims to transform noisy, unmasked motion into a complete, clean motion. 
\red{Recall that, in every reverse diffusion iteration of the motion completion task, the decoder replaces the unmasked motion latent with the original one after completion to keep the overall motion context.}
However, by incorporating both masked and unmasked motion into the iterative process, MMDM enables simultaneous refinement of the entire motion sequence. To achieve this, we fine-tune the model by updating both unmasked and masked motion using an updated loss function:
\begin{equation}
{\ell}_k^{\text{refine}} = \mathbb{E}_{k \sim \{1, K \}, \mathbf{d}_k} \big\Vert \mathbf{d}_k - \widehat{\mathbf{d}}_k \big\Vert_2. 
\end{equation}
Additionally, similar to previous works~\cite{jiang2024exploring,gong2023diffpose}, the reverse diffusion starts with the low-quality motion data instead of pure Gaussian noise. 
Other configurations remain consistent with the motion completion task.

\subsection{Motion In-betweening}\label{sec:mib}
Given a motion sequence $\textbf{d} = \{ \textbf{d}^{p}, \textbf{d}^{q}, \textbf{d}^{r} \}$ composed of a preceding part $\textbf{d}^{p} \in \mathbb{R}^{T_{0}\times J \times d}$, a transitioning part $\textbf{d}^{q} \in \mathbb{R}^{T_{1}\times J \times d}$, and a succeeding part $\textbf{d}^{r} \in \mathbb{R}^{T_{2}\times J \times d}$, we mask $\textbf{d}^{q}$ and encode $\textbf{d}^{p}$ and $\textbf{d}^{r}$, enabling the MMDM to reconstruct $\textbf{d}^{q}$. $T_0$, $T_1$, $T_2$ indicate the length for the preceding, transitioning, and succeeding parts, respectively.
In line with MDM~\cite{tevet2023mdm}, we employ a Transformer-encoder structure built by the Kinematic Encoder and integrate text embedding $\textbf{v}$ as an additional conditional embedding. 
Specifically, in each reverse diffusion iteration, $\textbf{d}^{p}$, $\textbf{d}^{q}_{k}$, $\textbf{d}^{r}$, and $\textbf{v}$ are concatenated, passed by a projection layer into the latent space. The Kinematic Encoder $E_{\phi}$ then processes it to yield the kinematic condition $c$, which is further transformed into motion signals by another linear layer to obtain $\textbf{d}^{q}_{k-1}$, completing the reverse iteration. The above process can be formulated as follows:
\begin{align}
\textbf{d}^{q}_{k-1} &= E_\phi(\mathbf{d}^{p} +\!\!\!+ \textbf{d}^{q}_{k} +\!\!\!+ \textbf{d}^{r} +\!\!\!+ \mathbf{v}, k).
\end{align}
More details can be found in the supplementary material.

{
\setlength{\tabcolsep}{0.5pt}
\begin{table}[t]
\caption{\red{Motion capture performance on Shelf and Campus~\cite{belagiannis2014multiple} datasets using the PCP (\%) metric. ``A-n'' corresponds to the $n$-th actor, while `AVG' denotes the average PCP. `$\dagger$' indicates the corrected value. \textbf{Bold} indicates the best performance; \underline{Underlined} indicates the second-best.}}
\vspace{-5pt}

\label{tab:shelf_campus}
\centering
\resizebox{0.46\textwidth}{!}{
\begin{tabular}{rcccccccc}
\toprule
\multirow{2}{*}{Method} & \multicolumn{4}{c}{Shelf} & \multicolumn{4}{c}{Campus} \\ 
\cmidrule(lr){2-5} \cmidrule(lr){6-9}
& A-1 & A-2 & A-3 & AVG & A-1 & A-2 & A-3 & AVG \\ 
\midrule
3DPS~\cite{belagiannis20153d} & 75.3$^{\;\;\;\;}$ & 69.7$^{\;\;\;\;}$ & 87.6$^{\;\;\;\;}$ & 77.5$^{\;\;\;\;}$ & 93.5$^{\;\;\;\;}$ & 75.7$^{\;\;\;\;}$ & 84.4$^{\;\;\;\;}$ & 84.5$^{\;\;\;\;}$ \\
4DAG~\cite{zhang20204d} & 99.0$^{\;\;\;\;}$ & 96.2$^{\;\;\;\;}$ & 97.6$^{\;\;\;\;}$ & 97.6$^{\;\;\;\;}$ & 64.8$^{\;\;\;\;}$ & 82.0$^{\;\;\;\;}$ & 96.6$^{\;\;\;\;}$ & 81.1$^{\;\;\;\;}$ \\
MVPose~\cite{dong2021fast} & 98.8$^{\;\;\;\;}$ & 94.1$^{\;\;\;\;}$ & 97.8$^{\;\;\;\;}$ & 96.9$^{\;\;\;\;}$ & 97.6$^{\;\;\;\;}$ & 93.3$^{\;\;\;\;}$ & 98.0$^{\;\;\;\;}$ & 96.3$^{\;\;\;\;}$ \\ 
TesseTrack~\cite{reddy2021tessetrack} & 99.1$^{\;\;\;\;}$ & 96.3$^{\;\;\;\;}$ & 98.3$^{\;\;\;\;}$ & 97.9$\dagger$$^{\;\;}$ & 97.9$^{\;\;\;\;}$ & 95.2$^{\;\;\;\;}$ & \textbf{99.1}$^{\;\;\;\;}$ & \underline{97.4}$^{\;\;\;\;}$ \\
PlaneSweep~\cite{lin2021multi} & 99.3$^{\;\;\;\;}$ & 96.5$^{\;\;\;\;}$ & 98.0$^{\;\;\;\;}$ & 97.9$^{\;\;\;\;}$ & \underline{98.4}$^{\;\;\;\;}$ & 93.7$^{\;\;\;\;}$ & \underline{99.0}$^{\;\;\;\;}$ & 97.0$^{\;\;\;\;}$ \\
D-MAE~\cite{jiang2022dual} & \textbf{99.7}$^{\;\;\;\;}$ & 94.1$^{\;\;\;\;}$ & \underline{98.4}$^{\;\;\;\;}$ & 97.4$^{\;\;\;\;}$ & 68.5$^{\;\;\;\;}$ & 90.3$^{\;\;\;\;}$ & 91.9$^{\;\;\;\;}$ & 83.6$^{\;\;\;\;}$ \\
QuickPose~\cite{zhou2022quickpose} & \underline{99.5}$^{\;\;\;\;}$ & 96.7$^{\;\;\;\;}$ & 98.2$^{\;\;\;\;}$ & 98.1$^{\;\;\;\;}$ & - & - & - & - \\
TEMPO~\cite{choudhury2023tempo} & 99.0$^{\;\;\;\;}$ & 96.3$^{\;\;\;\;}$ & 98.2$^{\;\;\;\;}$ & 97.8$\dagger$$^{\;\;}$ & 97.7$^{\;\;\;\;}$ &  \underline{95.5}$^{\;\;\;\;}$ & 97.9$^{\;\;\;\;}$ & 97.0$\dagger$$^{\;\;}$ \\
SelfPose3D~\cite{srivastav2024selfpose3d} & 97.2$^{\;\;\;\;}$ & 90.3$^{\;\;\;\;}$ & 97.9$^{\;\;\;\;}$ & 95.1$^{\;\;\;\;}$ & 92.5$^{\;\;\;\;}$ & 82.2$^{\;\;\;\;}$ & 89.2$^{\;\;\;\;}$ &  87.9$^{\;\;\;\;}$ \\
JCSAT~\cite{jiang2025JCSAT} & 99.3$^{\;\;\;\;}$ & \underline{97.0}$^{\;\;\;\;}$ & 98.2$^{\;\;\;\;}$ & \underline{98.2}$^{\;\;\;\;}$ & \textbf{99.6}$^{\;\;\;\;}$ & 93.5$^{\;\;\;\;}$ & 98.8$^{\;\;\;\;}$ & 97.3$^{\;\;\;\;}$ \\
\midrule
MMDM (Ours) & 98.4$^{\pm.08}$ & \textbf{97.1}$^{\pm.49}$ & \textbf{99.9}$^{\pm.05}$ & \textbf{98.5}$^{\pm.15}$ & \underline{98.4}$^{\pm.23}$ & \textbf{95.8}$^{\pm.08}$ & 98.5$^{\pm.12}$ & \textbf{97.6}$^{\pm.08}$ \\ 
\bottomrule
\end{tabular}}
\end{table}
}

\begin{figure*}[t]
\centering
\includegraphics[width=0.98\linewidth]{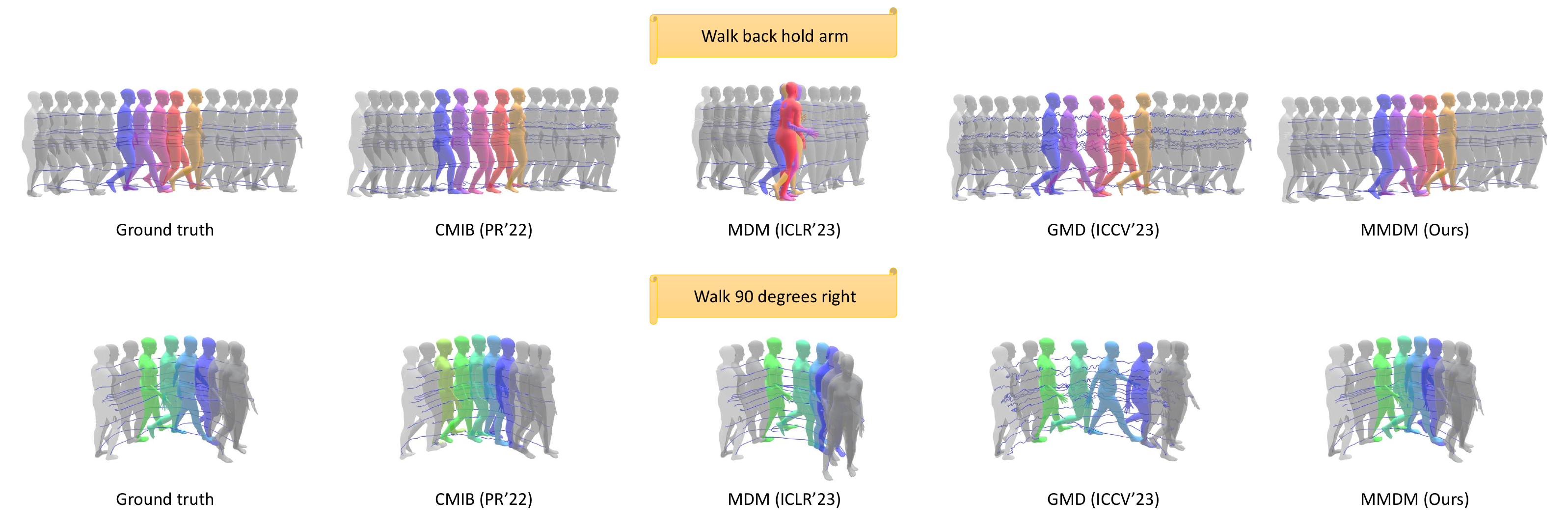}
\vspace{-13pt}
\caption{Qualitative comparisons for motion in-betweening task. Motion sequences are sampled into key poses at a fixed ratio. Grey segments illustrate preceding and succeeding parts, while the transitioning part is color-coded from yellow to purple in a rainbow gradient, indicating the chronological order. We emphasize the joint trajectories of the pelvis, elbows, shoulders, knees, and four end-effectors. Our model generates trajectories that are closest to the ground truth, whereas other methods suffer from issues such as over-smoothing and jitter.}
\label{fig:mib}
\end{figure*}

{
\setlength{\tabcolsep}{2.5pt}
\begin{table}[!t]
\begin{center}
\caption{\red{Motion capture performance on the BUMocap~\cite{jiang2022dual} (BU) and BUMocap-X~\cite{jiang2025JCSAT} (BU-X) datasets. Average PCP (\%), Precision (\%), Recall (\%), and MPJPE (mm) are reported.
}}
\vspace{-5pt}
\label{table:bumocap}
\resizebox{0.46\textwidth}{!}{
\begin{tabular}{rrcccc}
\toprule
Dataset & Method & PCP $\uparrow$ & Precision $\uparrow$ & Recall $\uparrow$ & MPJPE $\downarrow$ \\
\midrule
\multirow{4}{*}{BU} & 4DAG~\cite{zhang20204d} & 82.1$^{\;\;\;\;}$ & 95.1$^{\;\;\;\;}$ & 94.4$^{\;\;\;\;}$ & 70.8$^{\;\;\;\;}$ \\
& D-MAE~\cite{jiang2022dual} & 94.3$^{\;\;\;\;}$ & 97.5$^{\;\;\;\;}$ & 97.5$^{\;\;\;\;}$ & \underline{56.3}$^{\;\;\;\;}$ \\
& TEMPO~\cite{choudhury2023tempo} & 88.7$^{\;\;\;\;}$ & 86.4$^{\;\;\;\;}$ & 86.4$^{\;\;\;\;}$ & 96.5$^{\;\;\;\;}$ \\
& JCSAT~\cite{jiang2025JCSAT} & \underline{95.5}$^{\;\;\;\;}$ & 97.9$^{\;\;\;\;}$ & \underline{97.9}$^{\;\;\;\;}$ & \textbf{52.2}$^{\;\;\;\;}$ \\
\cmidrule(lr){2-6}
& MMDM (Ours) & \textbf{96.2}$^{\pm.52}$ & \textbf{98.1}$^{\pm.14}$ & \textbf{98.1}$^{\pm.14}$ & \textbf{52.2}$^{\pm.20}$ \\
\midrule
\multirow{4}{*}{BU-X} & 4DAG~\cite{zhang20204d} & 73.1$^{\;\;\;\;}$ & \underline{95.1}$^{\;\;\;\;}$ & \underline{95.1}$^{\;\;\;\;}$ & 56.5$^{\;\;\;\;}$ \\
& D-MAE~\cite{jiang2022dual} & 57.0$^{\;\;\;\;}$ & 77.5$^{\;\;\;\;}$ & 48.4$^{\;\;\;\;}$ & 113.4$^{\;\;\;\;}$ \\
& TEMPO~\cite{choudhury2023tempo} & 81.8$^{\;\;\;\;}$ & 77.1$^{\;\;\;\;}$ & 77.1$^{\;\;\;\;}$ & 136.5$^{\;\;\;\;}$ \\
& JCSAT~\cite{jiang2025JCSAT} & \underline{88.5}$^{\;\;\;\;}$ & \textbf{97.8}$^{\;\;\;\;}$ & \textbf{97.8}$^{\;\;\;\;}$ & \textbf{49.7}$^{\;\;\;\;}$ \\
\cmidrule(lr){2-6}
& MMDM (Ours) & \textbf{92.1}$^{\pm.43}$ & \underline{95.1}$^{\pm.89}$ & \underline{95.1}$^{\pm.89}$ & \underline{53.9}$^{\pm.62}$ \\
\bottomrule
\end{tabular}}
\end{center}
\end{table}
}

\section{Experiments}

\subsection{Implementation Details}

Our model is implemented \red{using PyTorch Lightning and optimized with AdamW, employing a learning rate of  $1\times10^{-5}$. 
Model training is performed on a workstation equipped with a single NVIDIA A100 GPU and an Intel Xeon Gold 6438Y+ CPU.}
\red{For the motion completion task, training is carried out over 150,000 iterations with a batch size of 128, requiring approximately 1.2 days to complete. For the motion refinement task, the model is further fine-tuned for an additional 50,000 iterations with the same batch size, which takes about 0.5 days.}
In both motion completion and refinement, OpenPose~\cite{cao2019openpose} is employed as the 2D pose estimator, and a skeleton transformation algorithm~\cite{zhang20204d} is applied to ensure consistent skeleton types across different datasets. The motion sequence length is set to 10 frames.
For the motion in-betweening task, 
\red{training is performed for 100,000 iterations with a batch size of 512 over one day.}
The transition length $T_1$ is set to 30 frames. To ensure fair comparison with existing methods, we adopt the same motion representation HumanML3D~\cite{guo2022generating}. We rearrange this representation to support joint-level encoding.

Further details (e.g., network structure, motion representations) are available in the supplementary material.

\subsection{Datasets and Metrics}

\noindent\textbf{Campus and Shelf.} These are two pioneering multi-view, multi-person mocap benchmarks established by Belagiannis et al.~\cite{belagiannis2014multiple}, with around 5000 frames in total. 
We train and evaluate our method on these two datasets for the motion completion and motion refinement tasks.
Following the common setting~\cite{belagiannis2014multiple,belagiannis20153d}, we exclude the 4th actor due to extensive occlusion, which results in unreliable ground truth.

\noindent\textbf{BUMocap and BUMocap-X.} BUMocap is an emerging multi-view multi-person mocap dataset introduced by D-MAE\cite{jiang2022dual}, along with its extended version, BUMocap-X~\cite{jiang2025JCSAT}. BUMocap includes 1000 frames capturing challenging 3-individual interactive motions from 5 cameras. BUMocap-X offers 3600 frames capturing 5-person group dance routines with serious occlusion. Both datasets provide precise annotations of 3D joint coordinates, as all data is manually annotated rather than automatically reconstructed. 
We utilize these two datasets for the motion completion task. The training and testing configurations are identical to previous studies~\cite{jiang2022dual,jiang2025JCSAT}.

\noindent\textbf{BABEL-TEACH.} This is a subset of BABEL~\cite{punnakkal2021babel}, rearranged by TEACH~\cite{athanasiou2022teach}, details motion sequences from the large mocap dataset AMASS~\cite{mahmood2019amass}, which includes around 10k motion sequences annotated with around 65k action labels. BABEL-TEACH provides distinct action segments, enabling us to focus on motion in-betweening for individual actions. Following TEACH, we removed ``T-pose'' and ``A-pose'' segments and ensured each segment contained at least 45 frames and no more than 224 frames. As a result, we have 10765 training segments and 1191 testing segments, totaling around 10.7 hours of motion data.

\noindent\textbf{Metrics.} 
For motion completion and motion refinement, we evaluate our model using standard metrics: 
\emph{1)} Percentage of Correctly estimated Parts (PCP) measures limb reconstruction accuracy; 
\emph{2)} Mean Per Joint Position Error (MPJPE) calculates the average Euclidean distance in millimeters between the predicted and the ground-truth joint; 
\emph{3)} Precision and recall, computing the ratio of correct joints among the estimated joints and ground-truth joints, with correctness defined as an Euclidean distance of less than 0.2 meters;
\red{\emph{4)} Mean Per Joint Acceleration Error (Accel) measures the average discrepancy in acceleration between the predicted and ground-truth joint motions, reflecting the smoothness and jitter error in millimeters per frame$^2$.}
For motion in-betweening, we assess the transition quality in accordance with prior works~\cite{harvey2020RMIB,kim2022CMIB}: 
\emph{1)} Normalized Power Spectrum Similarity (NPSS)~\cite{gopalakrishnan2019neural}, which measures the quality of motion trajectories in terms of the power spectrum of joint rotations; 
\emph{2)} two L2 norms of motion trajectories and joint rotations represented in quaternions, denoted as L2-P and L2-Q, respectively.

\subsection{Results and Analysis}

\red{To assess the stability and reliability of our method, we report the mean and standard deviation of each metric over 10 independent runs for all main experiments. In the following, we present the results for motion completion, motion refinement, and motion in-betweening, together with performance comparisons to state-of-the-art methods.}

\paragraph{Motion Completion} 
We integrated the MMDM into a mocap system and compared its capturing performance against state-of-the-art systems~\cite{choudhury2023tempo,jiang2025JCSAT,srivastav2024selfpose3d,zhang20204d} using the Shelf, Campus~\cite{belagiannis2014multiple}, BUMocap~\cite{jiang2022dual}, and BUMocap-X~\cite{jiang2025JCSAT} datasets.
Quantitative results are shown in Table~\ref{tab:shelf_campus} and Table~\ref{table:bumocap}, and qualitative comparisons are shown in Fig.~\ref{fig:result_shelf}. Table~\ref{tab:shelf_campus} indicates that our MMDM-integrated mocap system achieves the highest average PCP scores compared to others. 

Specifically, for the Shelf dataset, significant occlusion of the second actor causes other methods to fail, either missing joint positions entirely or reconstructing them inaccurately. In contrast, our method effectively completes these missing or manually masked joints, achieving the highest PCP score, highlighting the importance of motion completion and our approach's superiority.
For the Campus dataset, despite challenges like low resolution and occlusions causing poor 2D and 3D detections, our method notably improves the second actor's performance, achieving the highest average performance.

For the BUMocap and BUMocap-X datasets, Table~\ref{table:bumocap} reports the average PCP score, Precision, Recall, and MPJPE results. On BUMocap, our method outperforms existing approaches, including our previous work, JCSAT~\cite{jiang2025JCSAT}, across all metrics. 
\red{However, on BUMocap-X, our method achieves the highest score only for PCP, ranking second-best on the remaining metrics. This performance drop can be attributed to the severe occlusions present in BUMocap-X, which cause the simple mocap framework to estimate low-quality source motion. 
Unlike JCSAT, which addresses BUMocap-X by leveraging all triangulated joints from paired views to obtain denser inputs, our MMDM focuses on the completion of missing data and the enhancement of low-confidence results via manual masking, limiting modifications to masked or missing regions. Consequently, overall performance can be diminished when the input motion quality is poor. Nonetheless, our method consistently ranks within the top two among existing approaches, demonstrating its strong generalization and robustness across different datasets.
}

For qualitative results, Fig.~\ref{fig:result_shelf} illustrates the 2D projections and 3D renderings for the Shelf, BUMocap, and BUMocap-X datasets. We highlight failed reconstructions with red-dotted boxes. It shows that, unlike other methods that produce unnatural poses due to occlusions, MMDM outputs natural poses, highlighting its superiority. \red{More visual examples can be found in the supplementary material.}

{
\setlength{\tabcolsep}{1pt}
\begin{table}[!t]
\caption{\red{Quantitative comparisons for the motion refinement task on the Shelf dataset~\cite{belagiannis2014multiple}. We report the results before and after refinement, with the incremental change $\Delta$ (\%).} }
\vspace{-5pt}
\centering
\resizebox{0.46\textwidth}{!}{
\begin{tabular}{rcccccccccc}
\toprule
\multirow{2}{*}{Input} & \multirow{2}{*}{Method} & \multicolumn{3}{c}{PCP} & \multicolumn{3}{c}{MPJPE} & \multicolumn{3}{c}{Accel}  \\
\cmidrule(lr){3-5} \cmidrule(lr){6-8} \cmidrule(lr){9-11}
& & Before & After & $\Delta\uparrow$ & Before & After & $\Delta\uparrow$ & Before & After & $\Delta\uparrow$ \\
\midrule 
\multirow{4}{*}{\makecell{Noise\\(5 cm)}} & SmoothNet~\cite{zeng2022smoothnet} & \multirow{4}{*}{95.4} & 95.6 & 0.2 & \multirow{4}{*}{79.3} & 67.4 & 15.0 & \multirow{4}{*}{305.8} & 131.3 & \textbf{57.0} \\
& VPoser-t~\cite{rempe2021humor} & & 97.1 & 1.7 & & 70.9 & 10.5 & & 165.9 & 45.7 \\
& HuMoR~\cite{rempe2021humor} & & 97.8 & \underline{2.5} & & 63.4 & \underline{20.0} & & 134.4 & 56.0 \\
\cmidrule(lr){2-2} \cmidrule(lr){4-5} \cmidrule(lr){7-8} \cmidrule(lr){10-11}
& MMDM (Ours) & & 98.3$^{\pm .21}$ & \textbf{3.0} & & 60.2$^{\pm .31}$ & \textbf{24.0} & & 132.1$^{\pm .28}$ & \underline{56.8} \\
\midrule 
\multirow{4}{*}{\makecell{Noise\\(10 cm)}} & SmoothNet~\cite{zeng2022smoothnet} & \multirow{4}{*}{86.4} & 88.2 & 2.0 & \multirow{4}{*}{107.5} & 94.2 & 12.3 & \multirow{4}{*}{418.4} & 183.8 & \underline{56.0} \\
& VPoser-t~\cite{rempe2021humor} & & 88.1 & 1.9 & & 100.1 & 6.8 & & 271.5 & 35.1 \\
& HuMoR~\cite{rempe2021humor} & & 89.7 & \underline{3.8} & & 89.4 & \underline{16.8} & & 190.7 & 54.4\\
\cmidrule(lr){2-2} \cmidrule(lr){4-5} \cmidrule(lr){7-8} \cmidrule(lr){10-11}
& MMDM (Ours) & & 91.5$^{\pm .30}$ & \textbf{5.9} & & 84.2$^{\pm .32}$ & \textbf{21.6} & & 175.7$^{\pm .31}$ & \textbf{58.0} \\
\midrule 
\multirow{4}{*}{Mocap} & SmoothNet~\cite{zeng2022smoothnet} & \multirow{4}{*}{96.9} & 97.4 & 0.5 & \multirow{4}{*}{52.1} & 48.7 & 6.5 & \multirow{4}{*}{40.9} & 33.3 & \underline{18.5} \\
& VPoser-t~\cite{rempe2021humor} & & 97.4 & 0.5 & & 48.6 & 6.7 & & 38.0 & 7.0 \\
& HuMoR~\cite{rempe2021humor} & & 98.8 & \underline{1.9} & & 43.0 & \textbf{17.4} & & 35.0 & 14.4 \\
\cmidrule(lr){2-2} \cmidrule(lr){4-5} \cmidrule(lr){7-8} \cmidrule(lr){10-11}
& MMDM (Ours) & & 99.1$^{\pm .12}$ & \textbf{2.2} & & 43.1$^{\pm .14}$ & \underline{17.2} & & 32.0$^{\pm .11}$ & \textbf{21.7} \\
\bottomrule
\vspace{-5mm}
\end{tabular}}
\label{tab:shelf_campus_delta}
\end{table}
}

{
\setlength{\tabcolsep}{6pt}
\begin{table}[!t]
\caption{\red{Motion in-betweening results on the BABEL-TEACH~\cite{athanasiou2022teach} dataset. The performance of the 30-frame transition is reported. MDM$_{\star}$ and MDM$_{\star\star}$ represent two MDM variants.
$10\times$, $50\times$, $100\times$ denote results under DDIM sampling~\cite{song2021denoising}.
}}
\vspace{-5pt}
\centering
\begin{tabular}{rcccl}
\toprule
Method & L2-P $\downarrow$ & L2-Q $\downarrow$ & NPSS $\downarrow$ \\
\midrule
Interp. & \underline{0.0764}$^{\;\;\;\;}$ & \underline{0.0497}$^{\;\;\;\;}$ & 0.5501$^{\;\;\;\;}$ \\
RMIB~\cite{harvey2020RMIB} & 0.3830$^{\;\;\;\;}$ & 0.2345$^{\;\;\;\;}$ & 2.2503$^{\;\;\;\;}$ \\
CMIB~\cite{kim2022CMIB} & 0.0963$^{\;\;\;\;}$ & 0.0490$^{\;\;\;\;}$ & \underline{0.4687}$^{\;\;\;\;}$ \\
MDM~\cite{tevet2023mdm} & 0.2223$^{\pm.17}$ & 0.1419$^{\pm.09}$ & 1.1515$^{\pm.10}$ \\
GMD~\cite{karunratanakul2023gmd} & 0.2499$^{\pm.21}$ & 0.1109$^{\pm.18}$ & 1.0820$^{\pm.16}$ \\
\cmidrule(lr){1-4}
MDM$_\star$ & 0.2051$^{\pm.20}$ & 0.1361$^{\pm.18}$ & 1.2511$^{\pm.08}$ \\
MDM$_{\star\star}$ & 0.1800$^{\pm.15}$ & 0.1198$^{\pm.20}$ & 0.9100$^{\pm.11}$ \\
MMDM (Ours) & \textbf{0.0607}$^{\pm.15}$ & \textbf{0.0358}$^{\pm.19}$ & \textbf{0.2757}$^{\pm.13}$ \\
\bottomrule
\vspace{-5mm}
\end{tabular}
\label{tab:mib}
\end{table}
}

\paragraph{Motion Refinement} 
We have designed \red{three} experiments on the Shelf dataset to compare ours with three motion refinement methods: SmoothNet~\cite{zeng2022smoothnet}, VPoser-t~\cite{pavlakos2019expressive}, HuMoR~\cite{rempe2021humor}. \red{Performance was assessed using average PCP, MPJPE, Accel, and their respective incremental changes $\Delta = \frac{|\text{After} - \text{Before}|}{\text{Before}}$, with results summarized in Table~\ref{tab:shelf_campus_delta}. In the first two experiments, denoted as ``Noise (5 cm)'' and ``Noise (10 cm)'', Gaussian noise with standard deviations of 5 cm and 10 cm, respectively, was added to the 3D ground truth positions before refinement}. In the third experiment, labeled ``Mocap'', we applied these methods to refine motion data obtained from a real-world mocap system~\cite{dong2021fast}. 

As shown in Table~\ref{tab:shelf_campus_delta}, \red{our method outperforms competing approaches on most metrics, achieving the best results in 7 out of 9 cases and the second-best in the remaining 2. The cases where our method does not achieve the top score are the MPJPE metric in the ``Mocap'' experiment and the Accel metric in the ``Noise (5 cm)'' experiment; however, the differences from the best results in these cases are minimal.}

\red{
We attribute the slightly higher Accel but otherwise modest performance of SmoothNet to its tendency to reduce high-frequency motion components without contextual understanding, while HuMoR leverages kinematic cues to attain strong performance but, along with VPoser-t, only processes two frames at a time.
In contrast, our MMDM employs a sliding window to capture full temporal information. 
Moreover, MMDM demonstrates a deeper understanding of motion compared to HuMoR due to the KAA mechanism, which fuses joint- and pose-level representations and enhances the model's capacity to represent both skeletal structure and joint-level trajectories.
Overall, these results indicate that our model offers significant improvements, highlighting its advanced capability in understanding and completing diverse motion data.
}

\paragraph{Motion In-betweening}
Quantitative and qualitative results are presented in Table~\ref{tab:mib} and Fig.~\ref{fig:mib}. As shown in Table~\ref{tab:mib}, we include the interpolation result, labeled ``Interp.'' as the baseline, and compare our model with two motion in-betweening methods~\cite{harvey2020RMIB,kim2022CMIB} and two motion generation methods~\cite{tevet2023mdm,karunratanakul2023gmd}. Results demonstrate that our method outperforms others across all metrics, indicating that our approach can generate motion closest to the ground truth. Fig.~\ref{fig:mib} depicts preceding, transitioning, and succeeding motion segments for the ground truth and the generated motion by different methods. CMIB~\cite{kim2022CMIB} results in overly smooth key joint trajectories. MDM~\cite{tevet2023mdm} produces irrelevant poses and may alter the preceding and succeeding segments, which is undesirable. GMD~\cite{karunratanakul2023gmd}, due to the guidance from preceding and succeeding segments, results in minimal input changes, but leads to jitters. In contrast, our MMDM effectively generates appropriate transitioning segments, leveraging the motion understanding of Masked Autoencoders and the robust approximation capabilities of diffusion learning.

\subsection{Ablation Study}

To verify the effect of each proposed design, we mainly focus on two tasks: the motion refinement task on the Shelf dataset and the motion in-betweening task on the AMASS-TEACH testing dataset. 
\red{All tests are conducted on a single NVIDIA RTX 3090 GPU with an Intel i7-8700 3.20 GHz CPU.} 
More ablation studies for other tasks can be found in the supplementary material.

\paragraph{Motion Refinement}
We first investigate the impacts of the KAA, different masking training settings, denoising steps, and sampling steps on the motion refinement task.

\noindent\red{\textbf{Impact of Kinematic Attention Aggregation.}}
\red{We evaluate four encoder baselines: (1) a Transformer-based encoder attending solely to skeletal structure (``S''), (2) another attending only to key joint trajectories (``T''),  (3) a cascaded encoder comprising two Transformers that sequentially attend to structure and then trajectory (``S\&T''), and (4) a KAA variant that aggregates trajectory information before attending to skeletal structure (``KAA$\dagger$''). 
Table~\ref{tab:refine_ddim} presents the inference speed (FPS) and performance metrics (PCP, MPJPE, Accel) for our MMDM model and the listed baselines, where ``$1\times$'' denotes the final model incorporating the proposed KAA mechanism. Results show that combining structural and trajectory encoding improves performance, but significantly increases inference time. Both KAA and its variant—aggregating either skeletal structure first (see third row from bottom) or trajectory first (see fourth row from bottom)—further improve both accuracy and inference speed, attributable to KAA’s computational efficiency and its effective joint- and pose-level fusion. Importantly, aggregating skeletal structure first yields the highest accuracy, which motivates our choice of this configuration for the final KAA design. In summary, these results demonstrate the strong effectiveness of KAA.
}

\begin{figure}[!t]
\centering
\includegraphics[width=1.0\linewidth]{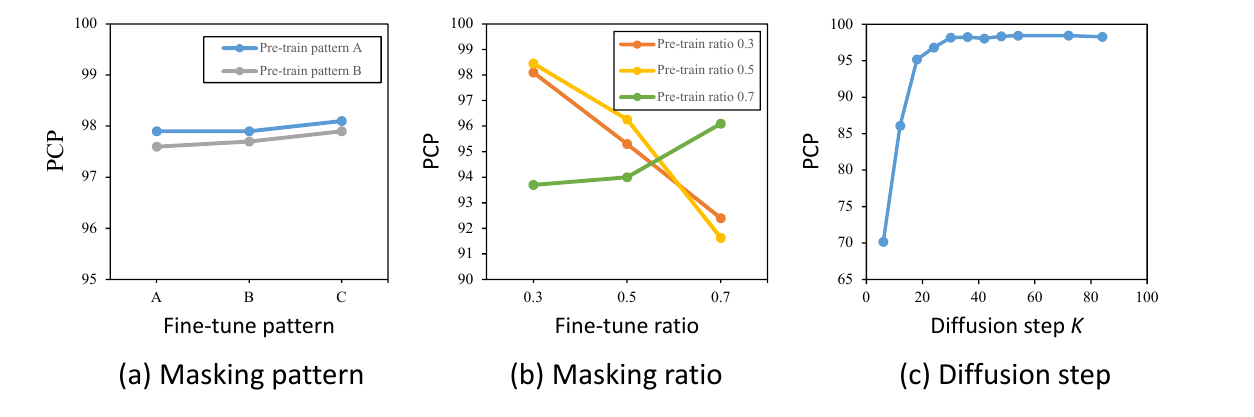}
\vspace{-25pt}
\caption{Ablation study for the motion refinement task. We report the average PCP on the Shelf dataset~\cite{belagiannis2014multiple}. The line graphs depict: \emph{(a)} horizontal axis for the fine-tune masking patterns, with lines showing pre-train masking patterns; \emph{(b)} horizontal axis for fine-tune masking ratios, with lines showing pre-train masking ratios; \emph{(c)} horizontal axis for diffusion steps.
}
\vspace{-5mm}
\label{fig:ab_refine}
\end{figure}

\noindent\textbf{Impact of Masking Patterns.}
We designed three masking patterns to select masked joints: \emph{a)} Pose-level random masking: joints from each pose are randomly masked at the same ratio, labeled ``Pattern A''; \emph{b)} Joint-level random masking: all joints across frames are masked at a masking ratio, labeled ``Pattern B''; \emph{c)} Weighted random masking: masking probability is calculated by Eq.~\ref{eq:fuse}, labeled ``Pattern C''. We first explored each pattern's effects to find the optimal one, using a 0.3 masking ratio. As shown in Fig.~\ref{fig:ab_refine}(a), pre-training the model with Pattern A and fine-tuning it with Pattern C yielded the best performance. This confirms the effectiveness of the proposed Adaptive Masking Process, enabling the model to treat key joints with low confidence or high triangulation error as noise and ignore them, thus capturing the accurate motion context.

{
\setlength{\tabcolsep}{6pt}
\begin{table}[!t]
\caption{\red{Motion refinements results on the Shelf dataset~\cite{belagiannis2014multiple} using ``Mocap'' data.
$5\times$ and $10\times$ denote the speed-up ratios under DDIM sampling~\cite{song2021denoising}.
}}
\vspace{-5pt}
\centering
\begin{tabular}{rrccccl}
\toprule
\multicolumn{2}{c}{Method} & PCP $\uparrow$ & MPJPE $\downarrow$ & Accel $\downarrow$ & FPS \\
\midrule
\multicolumn{2}{r}{SmoothNet~\cite{zeng2022smoothnet}} & {97.4}$^{\;\;\;\;}$ & {48.7}$^{\;\;\;\;}$ & 33.3 & 1257.2 \\
\multicolumn{2}{r}{HuMoR~\cite{rempe2021humor}} & \underline{98.8}$^{\;\;\;\;}$ & \textbf{43.0}$^{\;\;\;\;}$ & 35.0 & 31.9 \\
\midrule
\multirow{7}{*}{\rotatebox{90}{\makecell{MMDM\\(Ours)}}} & S & 98.2$^{\pm.08}$ & 49.2$^{\pm.11}$ & 35.6$^{\pm.15}$ & 154.1 \\
 & T & 98.3$^{\pm.11}$ & 49.0$^{\pm.13}$ & 35.1$^{\pm.14}$ & 148.2 \\
 & S\&T & 98.8$^{\pm.13}$ & 45.6$^{\pm.18}$ & 33.6$^{\pm.18}$ & 50.8 \\
 & KAA$\dagger$ & \underline{98.6}$^{\pm.25}$ & 44.9$^{\pm.19}$ & 33.1$^{\pm.23}$ &  107.9 \\
 & $1 \times$ & \textbf{99.1}$^{\pm.12}$ & \underline{43.1}$^{\pm.14}$ & \textbf{32.0}$^{\pm.11}$ & 108.4 \\
 \cmidrule(lr){2-6}
 & $5 \times$ & \textbf{99.1}$^{\pm.20}$ & 45.8$^{\pm.19}$ & \underline{32.4}$^{\pm.19}$ & 783.9\\
 & $10 \times$ & 98.7$^{\pm.24}$ & 47.9$^{\pm.21}$ & 33.5$^{\pm.15}$ & 1321.1\\
\bottomrule
\end{tabular}
\label{tab:refine_ddim}
\end{table}
}

\begin{figure}[!t]
\centering
\includegraphics[width=0.65\linewidth]{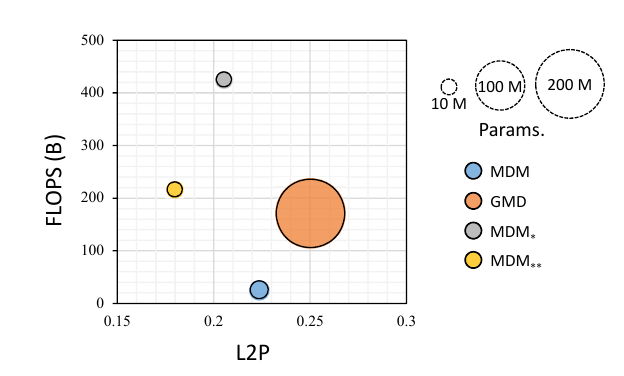}
\vspace{-10pt}
\caption{Computational complexity and motion in-betweening performance of each model are evaluated on the BABEL-TEACH dataset~\cite{athanasiou2022teach} using the L2P metric. The size of the circle indicates the number of parameters, with a larger area representing a greater number.}
\vspace{-3mm}
\label{fig:ab_flops}
\end{figure}

\noindent\textbf{Impact of Masking Ratios.}
We examined different masking ratios to optimize the model's comprehension. We tested ratios of $\{ 0.3, 0.5, 0.7 \}$ during both the pre-training and fine-tuning stages, using Pattern A as the default. As shown in Fig.~\ref{fig:ab_refine}(b), a high masking ratio during pre-training hinders effective motion reconstruction learning. The optimal model performance is observed with a 0.5 masking ratio for pre-training and a 0.3 masking ratio for fine-tuning.

\noindent\textbf{Impact of Denoising Step Numbers.} 
We treat input motion sequences as intermediate results, so refining motion at the beginning of the reverse diffusion process is unnecessary. We experimented with different diffusion steps to find an appropriate number.
As shown in Fig.~\ref{fig:ab_refine}(c), the PCP value increases and stabilizes when the model is configured with $K=50$. Consequently, we selected 50 as the number of diffusion steps.

\noindent\textbf{\red{Sampling Strategies.}}
\red{To accelerate the denoising process, a common way is to apply the DDIM sampling strategy~\cite{song2021denoising}. To analyze the acceleration impact, we conduct several experiments with different speed-up ratios ($N\times$). Here we use the setting with the total number of the reverse diffusion step 50, and $1\times$ stands for the baseline which utilizes normal sampling strategy DDPM~\cite{ho2020denoising}, and others stands for the DDIM sampling strategy. As shown in the last two rows of Table~\ref{tab:refine_ddim}, our model with DDIM acceleration achieves competitive inference speed relative to current methods, while delivering superior performance in both PCP and Accel metrics. Moreover, even without DDIM acceleration, our model maintains an FPS above 100, which satisfies most real-time requirements.}

\paragraph{Motion In-betweening} We now examine the effects of the joint-level representation, KAA, and sampling strategy on the motion in-betweening task.

\noindent\textbf{Impact of Joint-level Representation.}
Aforementioned, motion generation models such as \cite{tevet2023mdm,karunratanakul2023gmd} commonly adopt pose-level representation for human motion modeling. However, joint-level representation offers a more thorough exploration of temporal information across frames, as demonstrated in many HPE studies~\cite{zhang2022mixste,li2022exploiting}. Thus, we now verify the merit of employing such representation in generation models. We first developed a variant of the MDM model that includes both a Spatial encoder and a Temporal encoder. This variant has fewer parameters than the original model, and its motion in-betweening performance is reported in the 6th row of Table~\ref{tab:mib}, labeled as MDM$_{\star}$. Compared to the original model (3rd row), this variant shows an 8\%  improvement in L2-P and 5\% improvement in L2-Q, highlighting the potential of joint-level representation.

\noindent\textbf{Impact of Kinematic Attention Aggregation.} 
Next, we developed a second variant of the MDM model using the proposed Kinematic Encoder. This variant has the same number of parameters as the first variant, and its performance is reported in the 7th row of Table~\ref{tab:mib}, labeled as MDM$_{\star\star}$. Compared to the original one, this variant shows a substantial improvement, with a 19.5\% increase in L2-P, a 15.6\% increase in L2-Q, and a 20.9\% increase in NPSS. 
Moreover, as shown in Fig.~\ref{fig:ab_flops}, we present each model's performance using the L2P metric, with its computational complexity in billions of Floating Point Operations Per Second (FLOPS), and parameter count in millions. The figure reveals that joint-level representation enhances MDM's performance but increases its computational complexity by over 40 times. In contrast, the proposed KAA not only improves MDM's performance but also keeps its computational complexity within an acceptable range.
Overall, this suggests that our aggregation approach efficiently extracts motion context and facilitates a thorough exchange among spatial and temporal domains. 

\section{Limitations and Future Plans}

While MMDM demonstrates strong performance across various tasks, one limitation lies in its reliance on the computationally intensive reverse diffusion process. For example, generating motion sequences for in-betweening requires up to 1000 iterations, with evaluation times exceeding seven hours for a single variant. Future work will explore acceleration strategies such as DDIM~\cite{song2021denoising} and Consistency Models~\cite{song2023consistencymodels} to improve efficiency.
Another limitation is that while certain tasks can benefit from a shared pretraining stage, tasks with fundamentally different input-output characteristics, such as motion refinement compared to motion in-betweening, often require dedicated retraining or specialized adaptation. This divergence limits the applicability of a single model across all tasks without further adjustment. In future work, we aim to more fully exploit the transferable motion prior to enable a unified model that captures shared motion structures while allowing efficient fine-tuning for tasks with divergent requirements.

\section{Conclusion}

In this work, we introduce the Masked Motion Diffusion Model (MMDM), a diffusion-based generative framework designed to enhance vision-based motion capture systems by reconstructing missing or low-confidence-filtered motion data. MMDM is built upon a Masked Autoencoder structure and integrates the proposed Kinematic Attention Aggregation (KAA) mechanism, which enables efficient, deep, and iterative encoding of structural and temporal features by fusing joint-level and pose-level motion cues. A key contribution of our approach is the use of a single, reusable architecture capable of learning context-adaptive motion priors—task-specific representations that emphasize different aspects of motion dynamics. This adaptability allows the same model structure to specialize for diverse tasks such as motion completion, refinement, and in-betweening without architectural changes. MMDM achieves state-of-the-art results across multiple motion capture benchmarks, with ablation studies confirming the contribution of each component to the model’s effectiveness.

\section*{Acknowledgments}
This research was supported by the Theme-based Research Scheme, Research Grants Council of Hong Kong (T45-205/21-N), and the Guangdong and Hong Kong Universities “1+1+1” Joint Research Collaboration Scheme (2025A0505000003).

\bibliographystyle{IEEEtran}

\bibliography{myRefs}


\clearpage

\twocolumn[
    \begin{@twocolumnfalse}
        \customtitle
    \end{@twocolumnfalse}
]
\setcounter{section}{0}
\setcounter{figure}{0}
\setcounter{table}{0}

\title{Supplementary Material for MMDM}

\markboth{Manuscript submitted to IEEE Transactions on Multimedia}%
{Shell \MakeLowercase{\textit{et al.}}: A Sample Article Using IEEEtran.cls for IEEE Journals}

\maketitle

\section{Motion completion}

We employ the proposed Masked Motion Diffusion Model (MMDM) to a motion capture framework, as a post-processing step to fill the missing or manually masked motion sequences (we name it completion), thereby enhancing overall capture performance. We will now delve into the specifics, including training details, network structures, and more.

\subsection{Data preparation}\label{sec:data_prepare}

For the preparation of training and testing data, we incorporate normalization and augmentation and report their effects as follows. \textit{(a) Normalization:} We normalize each frame's 3D pose by subtracting the pose centroid, calculated by averaging the positions of key joints. As shown in Table~\ref{table:data_prepare}, using the hip position for subtraction results in the average PCP value dropping to 93.9 from 98.5 on the Shelf dataset.
\textit{(b) Augmentation:} Following normalization, we apply two augmentation methods to improve the model's generalizability: \emph{1)} Randomly rotating the 3D pose along the yaw axis within a range of -180 to 180 degrees; \emph{2)} Flipping the positions of the right and left body parts. In contrast to VideoPose3D~\cite{pavllo20193d}, we do not apply flip augmentation during the testing time, in order to avoid increased computational complexity. As shown in Table~\ref{table:data_prepare}, when we remove the rotation augmentation, the average PCP value drops to 97.1 from 98.5 on the Shelf dataset. When we remove the flipping augmentation, the average PCP value drops to 96.4 from 98.5 on the same dataset.

\subsection{Network structure}

Our MMDM employs a Masked Autoencoder framework to effectively encode the motion context, which is then used to guide the subsequent reverse diffusion process for the motion completion task. 
The model's architecture follows an Autoencoder structure, consisting of a kinematic encoder, featuring the proposed Kinematic Attention Aggregation (KAA) mechanism, and a conventional Transformer-based decoder. Both components incorporate standard Attention modules~\cite{vaswani2017attention}, including Multiheaded Self-Attention blocks (MSAs), Multiheaded Cross-Attention blocks (MCAs) and Feed-Forward Network blocks (FFNs), with Layernorm (LN) applied before each module. Specifics of the MMDM's architecture are outlined in Table~\ref{table:jcsat}. ``Head'' and ``Head Dim.'' denote the number of heads and the dimension of the feature used in MSAs and MCAs, respectively. We use a linear layer to adjust the feature to achieve a consistent dimension between MSAs and MCAs. \textit{The complete code of our framework will be released after publication.}

\begin{table}[t]
\caption{Impact of combinations of the data preparation processes. ``Rot.'' denotes the random rotation augmentation. We provide the average PCP (Avg. PCP) (\%) for motion completion conducted on the Shelf dataset.}
\centering
\label{table:data_prepare}
\begin{tabular}{ccccc}
\toprule
\multicolumn{2}{c}{Normalization} & \multicolumn{2}{c}{Augmentation} & \multirow{2}{*}{Avg. PCP $\uparrow$} \\ 
\cmidrule(l){1-2} \cmidrule(l){3-4}
Hip & Centroid & Rot. & Flip & \\
\midrule
$\checkmark$ & & & & 94.9 \\
$\checkmark$ & & $\checkmark$ & & 95.7 \\
$\checkmark$ & & & $\checkmark$ & 96.1 \\
$\checkmark$ & & $\checkmark$ & $\checkmark$ & 97.8 \\
 & $\checkmark$ & & & 96.2 \\
 & $\checkmark$ & $\checkmark$ & & 96.4 \\
 & $\checkmark$ & & $\checkmark$ & 97.1 \\
 & $\checkmark$ & $\checkmark$ & $\checkmark$ & 98.5 \\
\bottomrule
\end{tabular}
\end{table}

\setlength{\tabcolsep}{4pt}
\begin{table}[t]
\caption{The detailed structure of the Masked Motion Diffusion Model, for motion completion and motion refinement tasks.}
\centering
\label{table:jcsat}
\begin{tabular}{lclc}
\toprule
Encoder & Num. & Decoder & Num. \\ 
\cmidrule(l){1-2} \cmidrule(l){3-4}
In Dim. & 3 & In Dim. & 512 \\
Feat. Dim. & 512 & Feat. Dim. & 512 \\
Depth & 9 & Depth & 3 \\
Head & 4 & Head & 4 \\
Head Dim. & 32 & Head Dim. & 32 \\
FFN Dim. & 512 & FFN Dim. & 512 \\
Out Dim. & 512 & Out Dim. & 3 \\
\bottomrule
\end{tabular}
\end{table}

\subsection{Additional qualitative results}

\paragraph{Video-based qualitative comparisons} To showcase the smoothness and naturalness of the completed motion, we have included additional video results on the Shelf dataset. We compared our MMDM with other state-of-the-art motion capture  frameworks~\cite{zhang20204d,choudhury2023tempo,jiang2025JCSAT}. 
These visual examples are composed into a side-by-side comparison video. In this video, we provide the projected motion data illustrated with blue lines and the motion completion results in red lines for our method.
This side-by-side comparison demonstrates that, with the help of MMDM, missing joint positions are accurately completed, and low-confidence joint positions are effectively replaced with high-quality ones. Therefore, more fidelity mocap results are produced. 

\begin{figure}[t]
\centering
\includegraphics[width=0.95\linewidth]{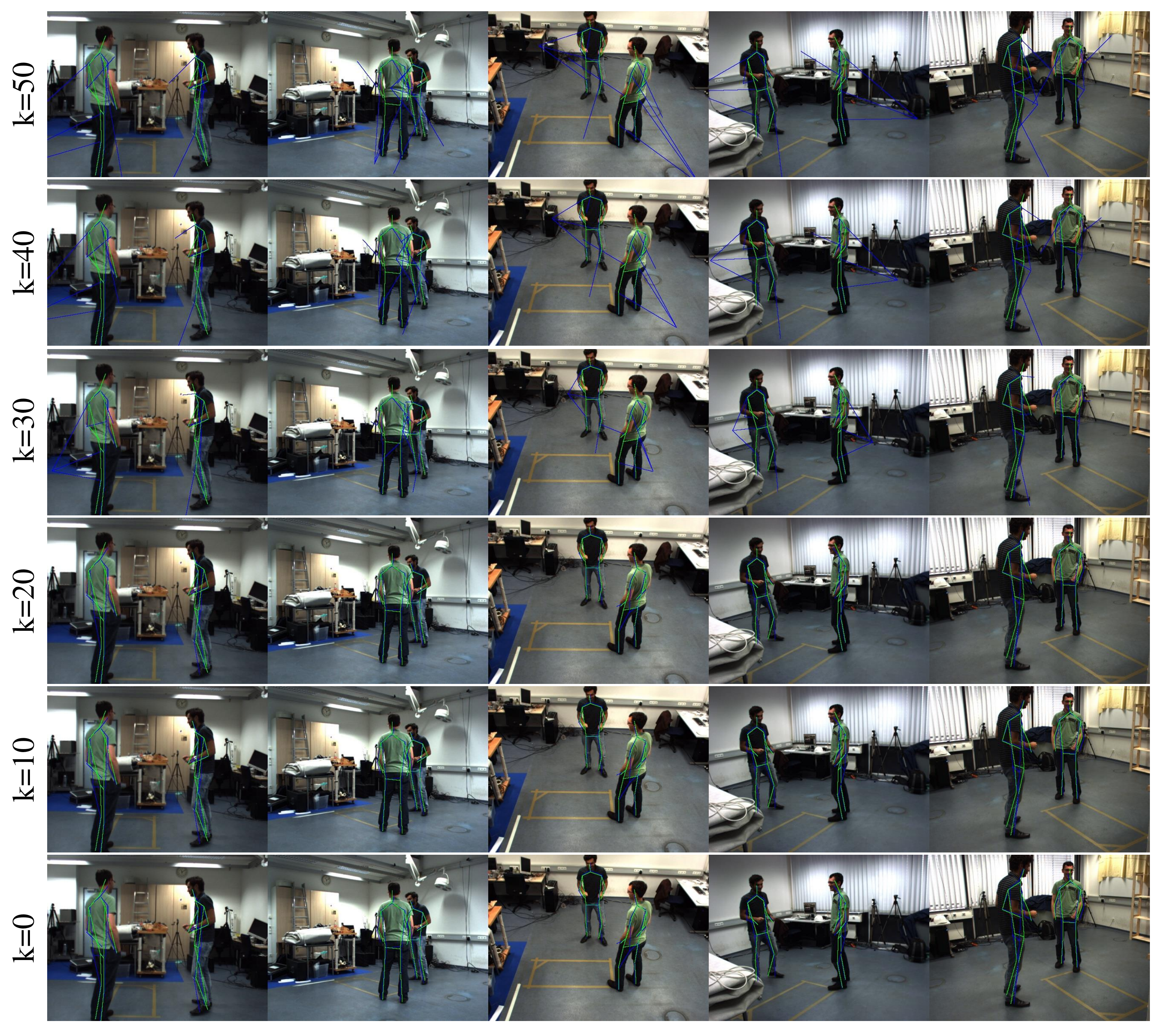}
\caption{Demonstration for the reserve diffusion process at $k$ time step. Green and blue skeletons denote the ground truth and the prediction, respectively. The masked joints are first sampled from a normal distribution and iteratively denoised.}
\label{fig:diff}
\end{figure}

\paragraph{Visualization of reserve diffusion} Fig.~\ref{fig:diff} depicts the reserve diffusion process on the Shelf dataset. The green skeletons represent the ground truth, while the blue skeletons are the outputs of the MMDM. Each column displays 2D projections from all views, with intermediate results shown from top to bottom at diffusion steps $k=\{50, 40, 30, 20, 10, 0\}$. During each diffusion step, all unmasked key joints are retained, and only the masked key joints (four masked in total) are denoised/updated.
Fig.~\ref{fig:diff} demonstrates that, as the diffusion step $k$ decreases sequentially, the masked parts transition from random noise to meaningful signals, forming accurate human poses. The proposed Kinematic Encoder effectively extracts the appropriate motion context. Additionally, in the early steps (from 50 to 30), the unmasked key joints quickly transit from noise, while changes become less noticeable in the later steps (from 20 to 0). This shows that this reverse diffusion process rapidly fits the ground truth in the early stages and makes fine adjustments later.

\subsection{Additional ablation studies}

\paragraph{Impact of data pre-processing} As mentioned, we adopt normalization and augmentation in the training data. We report the results in Table~\ref{table:data_prepare} with further analysis in Sec.~\ref{sec:data_prepare}.

\paragraph{Diffusion objective}
We evaluate the completion performance using different diffusion objectives on the Shelf dataset: to predict noise or to predict signal. Table~\ref{table:ab_mc} presents the average PCP scores when: \emph{1)} the diffusion model predicts the noise, denoted as $\epsilon$; \emph{2)} the diffusion model predicts the signal, denoted as $\mathbf{d}_0$. The 1st and 2nd rows of the table reveal that, there are minimal changes when switching the objective from signal to noise. Consequently, in the main paper, we choose to use the signal objective.

\paragraph{Impact of model size}
To find the optimal structure design, we created three variants of the MMDM and evaluated them on the Shelf dataset using the average PCP metric: \emph{1)} Lite version. MMDM's encoding and decoding feature dimensions are decreased from 512 to 256. The depth of the Encoder is decreased from 9 to 6; \emph{2)} Small version. MMDM's encoding and decoding feature dimensions are decreased from 512 to 256. \emph{3)} Large vision. The depth of the Encoder is increased from 9 to 12. As shown in Table~\ref{table:ab_mc}, ``Normal'' represents the original version. From the 2nd to the 5th row we can see that, as the network size increases and its depth becomes deeper, the performance initially improves but then shows no significant further improvement. This indicates that, the normal version is the optimal network structure design, as larger models do not provide substantial performance gains.

\paragraph{\red{Impact of 2D Pose Estimators}}
\red{To further evaluate the generalization and stability of our model when presented with 2D pose inputs produced by different 2D pose estimators, we conduct additional experiments using results from two recent estimators in addition to OpenPose~\cite{cao2019openpose}: AlphaPose~\cite{fang2022alphapose} and SimCC~\cite{li2022simcc}. 
Recall that OpenPose is a bottom-up framework optimized for real-time inference, computing Part Affinity Fields to associate body parts. In contrast, AlphaPose is a top-down framework that first detects target bodies and then estimates the pose within each cropped region, achieving higher accuracy. SimCC adopts a coordinate classification approach, encoding continuous coordinates as discrete probability distributions, which increases robustness against occlusions and detection ambiguities.}
\red{To unify the input formats, we apply the skeleton conversion algorithm~\cite{zhang20204d} to align the skeleton definitions across estimators. All testing protocols remain consistent with the experimental settings described in the main paper; notably, the model used for the motion completion task is evaluated without additional training or fine-tuning. Following the evaluation protocol, we report average PCP and MPJPE on the Shelf dataset. As shown in Table~\ref{table:ab_2d_pose}, results obtained using different 2D pose estimators exhibit only minor variations. This indicates that training the model with OpenPose outputs and testing it with other estimators does not introduce significant discrepancies or data shift, thereby demonstrating the strong generalization and stability of our MMDM framework with respect to diverse 2D pose inputs.}

\setlength{\tabcolsep}{8pt}
\begin{table}[t]
\caption{Motion completion performance on the Shelf dataset, using the average PCP metric. We evaluate the MMDM under different objective and network structure settings.}
\centering
\label{table:ab_mc}
\begin{tabular}{cccc}
\toprule
\multicolumn{2}{c}{Objective} & \multirow{2}{*}{Version} & \multirow{2}{*}{Avg. PCP $\uparrow$} \\ 
\cmidrule(l){1-2}
$\epsilon$ & $\mathbf{d}_0$ & & \\
\midrule
\checkmark & & Normal & 98.47 \\
& \checkmark & Normal & \underline{98.48} \\
& \checkmark & Lite & 95.41 \\
& \checkmark & Small & 96.90 \\
& \checkmark & Large & \textbf{98.52} \\
\bottomrule
\end{tabular}
\end{table}

{
\setlength{\tabcolsep}{4pt}
\begin{table}[t]
\caption{\red{Motion completion performance on the Shelf dataset, using the average PCP and MPJPE metrics. We evaluate the MMDM under different 2D pose inputs.}}
\centering
\label{table:ab_2d_pose}
\begin{tabular}{rccc}
\toprule
Estimator & Avg. PCP $\uparrow$ & MPJPE $\downarrow$ & Pub. Venue \\ 
\midrule
AlphaPose~\cite{fang2022alphapose} & \underline{98.5} & \underline{44.7} & TPAMI'22 \\
SimCC~\cite{li2022simcc} & \textbf{98.6} & \textbf{44.8} & ECCV'22 \\
\midrule
OpenPose~\cite{cao2019openpose} & \underline{98.5} & 45.1 & TPAMI'19\\
\bottomrule
\end{tabular}
\end{table}
}

\section{Motion refinement}
For motion completion, we only denoise/update the masked key joints through the diffusion process. For motion refinement, we denoise/update the whole sequence. The network structure remains the same for both tasks.

\subsection{Additional qualitative comparisons} To demonstrate the inter-frame continuity of the refined movements, we have provided three video clips that have been composed into one side-by-side comparison video. We apply the proposed MMDM into the existing motion capture framework and utilize it to refine the motion data. These clips showcase the performance of various motion capture systems on the Shelf dataset. Within these clips, green represents the ground truth, light yellow indicates the original input, and blue signifies the refined results from our MMDM. When examining the original motion (within light yellow colour), MVPose~\cite{dong2021fast} has the poorest inter-frame continuity, evidenced by the jittering of most joints. 4DA~\cite{zhang20204d} has the highest inter-frame continuity, owing to the application of a temporal filter. However, the temporal filter has caused a noticeable misalignment in the lower limbs of the actor who wears a green T-shirt. Our refined results (within blue colour) show significant improvement over the original input, despite a minor jitter in key joints.

\begin{algorithm}[t]
\caption{Motion in-betweening via the reverse diffusion}
\label{alg:cap}
\begin{algorithmic}[1]
\Require A preceding motion segment $\textbf{d}^{m}$, a succeeding motion segment $\textbf{d}^{l}$, a diffusion model $G_{\varphi}$, a goal function $F (\cdot, \cdot)$ (if any), an Emphasis matrix $M$ (if any) and its inverse matrix $M^{-1}$, a scaling factor $s$ of Dense Gradient Propagation (if any).

\State $\textbf{d}_{K} \gets$ sample from $\mathcal{N}(0,\mathbf{I})$
\State $\textbf{d}^{m} \gets \textbf{d}^{m} \cdot M $ \Comment{If Emphasis Projection enables}
\State $\textbf{d}^{l} \gets \textbf{d}^{l} \cdot M $ \Comment{If Emphasis Projection enables}

\ForAll{ $k$ from $K$ to $1$}
\State $\textbf{d}_{0} \gets G_{\varphi}(\textbf{d}_{k})$
\State $\textbf{d}_{0}^{\prime} \gets$ Concatenate $(\textbf{d}^{m},\textbf{d}_{0}^{n},\textbf{d}^{l})$ \Comment{Imputation on the preceding and succeeding segments}
\State $\mu, \Sigma \gets \mu(\textbf{d}_{0}^{\prime}, \textbf{d}_{k}),\Sigma_t$
\State $\Delta \gets -s \sum \nabla_{\textbf{d}_{k}}(F(\textbf{d}^{m},\textbf{d}_{0}^{m}) +F(\textbf{d}^{l},\textbf{d}_{0}^{l}))   $ \Comment{If Dense Gradient Propagation enables}
\State $\mu \gets \mu + \Delta$
\State $\textbf{d}_{k-1} \sim \mathcal{N}(\mu,\Sigma)$
\EndFor
\State $\textbf{d}_{0} \gets \textbf{d}_{0} \cdot M^{-1}$
\State \Return $\textbf{d}_{0}$
\end{algorithmic}
\end{algorithm}

\section{Motion in-betweening}

Give a motion sequence $\textbf{d}=\{d_{t}\}_{t=1}^{T} \in \mathbb{R}^{T\times J\times d}$, where $T$ denotes the frame number and $J$ denote the key joint number per frame, each with a representation dimension $d$. We split it into three segments: the preceding segment $\textbf{d}^{p} \in \mathbb{R}^{T_{0}\times J\times d}$, the transitioning segment $\textbf{d}^{q} \in \mathbb{R}^{T_{1}\times J\times d}$, and the succeeding segment $\textbf{d}^{r} \in \mathbb{R}^{T_{2}\times J\times d}$, where $T=T_{0}+T_{1}+T_{2}$ and $T_1 = 30$. The objective here is to generate $\textbf{d}^{q}$ under the conditions of $\textbf{d}^{p}$, $\textbf{d}^{r}$ and this sequence's CLIP~\cite{radford2021learning}-encoded action label $\textbf{v}$.

\subsection{Joint-level motion representation}

We use HumanML3D~\cite{guo2022generating} as the motion representation, following the skeleton structure of SMPL~\cite{loper2015smpl}, where the total number of joints $J=22$. Sequentially,  $\textbf{d}$ is defined as a series of tuples $d_{t} = ({r}^{v},\dot{r}^{a},\dot{r}^{x},\dot{r}^{z},{r}^{y},\textbf{j}^p,\textbf{j}^v,\textbf{j}^r,\textbf{c}^f)$. Here, $\dot{r}^{a} \in \mathbb{R}^1$ represents the root angular velocity along the pitch axis; $\dot{r}^{x} \in \mathbb{R}^1$, $\dot{r}^{z} \in \mathbb{R}^1$ are the root linear velocities along the yaw axis; $r^{y} \in \mathbb{R}^1$ is the root height; $\textbf{j}^p \in \mathbb{R}^{J \times 3}$, $\textbf{j}^v \in \mathbb{R}^{J \times 3}$, $\textbf{j}^r \in \mathbb{R}^{(J-1) \times 6}$ are the local joint positions, linear velocities and rotations in Rotation6D format~\cite{zhou2019continuity}; and $\textbf{c}^f \in \mathbb{R}^4$ denotes the binary foot contact labels. Additionally, to facilitate converting this representation back to SMPL‐based motion, the root rotation velocity ${r}^{v} \in \mathbb{R}^6$ is included, resulting in a final motion representation $\textbf{d} \in \mathbb{R}^{T \times 269}$.
However, our model requires joint‐level motion as input, whereas the current motion representation is at the pose level. To address this, we rearrange $\textbf{d}$ by concatenating $(\dot{r}^{a},\dot{r}^{x},\dot{r}^{z},{r}^{y},\textbf{c}^f)$ and $({r}^{v}, \textbf{j}^r,\textbf{j}^p,\textbf{j}^v)$ to form a joint-level representation $\textbf{d}^{\prime} \in \mathbb{R}^{T\times J^{\prime} \times 12}$, where $J^{\prime} = 22$.

\setlength{\tabcolsep}{8pt}
\begin{table}[t]
\caption{The detailed structure of the proposed Masked Motion Diffusion Model, for motion in-betweening.}
\centering
\label{table:mib_structure}
\begin{tabular}{lc}
\toprule
Encoder & Num. \\ 
\midrule
In Dim. & 12 \\
Feat. Dim. & 512 \\
Depth & 8 \\
Head & 8 \\
Head Dim. & 64 \\
FFN Dim. & 1024 \\
Out Dim. & 12 \\
\bottomrule
\end{tabular}
\end{table}

\subsection{Network structure and implementation details}

Following MDM~\cite{tevet2023mdm}, we adopt a Transformer encoder structure and utilize two linear layers to project the motion from the signal to the latent space and then back. Table~\ref{table:mib_structure} demonstrates the specifics of our MMDM's architecture.
We implement the MMDM, MDM~\cite{tevet2023mdm} and GMD~\cite{karunratanakul2023gmd} using Pytorch Lightning~\cite{pytorch_lightning} for network construction and Hugging Face's Diffusers~\cite{huggingface} for the diffusion pipeline implementation. All re-implementations strictly adhere to the hyperparameters specified in the original papers. Following \cite{tevet2023mdm,karunratanakul2023gmd}, we utilized a frozen CLIP-ViT-B/32 to encode action labels.

\subsection{Inference details}

\paragraph{Motion imputation}
To enable the motion diffusion models to generate the transitioning segment, MDM~\cite{tevet2023mdm} suggests to overwrite the generated $\widehat{\textbf{d}^{p}}$ and $\widehat{\textbf{d}^{r}}$ with the input ${\textbf{d}^{p}}$ and ${\textbf{d}^{r}}$ during the reverse diffusion process. MDM believes this overwrite process helps ensure that the generated output stays consistent with the original inputs (preceding and succeeding) while filling in the missing part (transitioning). GMD~\cite{karunratanakul2023gmd} names it as motion imputation and introduces two strategies (\textit{Emphasis Projection} and \textit{Dense Gradient Propagation}, as detailed in the next section) to enhance the guidance from the inputs on the missing part. We follow \cite{tevet2023mdm,karunratanakul2023gmd}'s motion imputation process, which can be summarized in Algorithm~\ref{alg:cap}.

\paragraph{GMD's pre-processing}

GMD~\cite{karunratanakul2023gmd} introduces two strategies to improve guided motion generation performance: 1) \textit{Emphasis Projection}. It increases the relative importance of the root trajectory compared to other joint movements in each frame. This root importance is parameterized by an Emphasis matrix $M$, controlled by the emphasis factor; 2) \textit{Dense Gradient Propagation}. It calculates the gradient of the distance between the target and generated key joint positions during each iteration of the reverse diffusion process, with a scaling factor $s$ controlling its influence on the generated motion. 
In our main paper, we implement these pre-processing steps for GMD~\cite{karunratanakul2023gmd} and present the resulting motion in-betweening performance in Table~\ref{tab:mib_ab}. For Emphasis Projection, we set the emphasis factor to 10, and due to GPU resource constraints, we only assess its effectiveness by enabling or disabling it, without exploring variations in its value. For Dense Gradient Propagation, we use the L2-norm as the goal function, comparing the generated preceding and succeeding segments with the ground truth to apply the guidance from these segments to the transitioning segments. Evaluation of the scaling factor is detailed in the next section.

\subsection{Additional ablation studies}

\paragraph{Impact of GMD's pre-processing}

As shown in Table~\ref{tab:mib_ab}, ``Rand. Proj.'' refers to the Emphasis projection, and ``Scale'' indicates the influence intensity of the dense gradient propagation.

\noindent\textbf{Emphasis Projection.}
We applied the Emphasis Projection to both MDM~\cite{tevet2023mdm} and our MMDM to evaluate its impact. From the 2nd, 3rd, 9th and 10th rows, we observe that this technique improves performance in some metrics. However, for MMDM, only one out of three metrics showed improvement, indicating that this technique may not significantly enhance MMDM performance. Therefore, in the main paper, we did not include the results of MMDM using Emphasis Projection.

\noindent\textbf{Scaling of Dense Gradient Propagation.} 
From the 4th to 8th rows of Table~\ref{tab:mib_ab}, we observe that as the scaling factor increases, the overall performance initially improves and then declines. The best performance is observed in the 7th row. We believe this is because the guidance introduces jitters, reducing the smoothness of the generated motion.
This also indicates that for any motion, achieving a smooth and natural motion in-betweening result with GMD requires carefully adjusting the scaling factor, which presents a challenge for optimization.

\setlength{\tabcolsep}{3.pt}
\begin{table}[!t]
\caption{Impact of GMD~\cite{karunratanakul2023gmd}'s pre-processing in the motion in-betweening task. The performance of the 30-frame transition on the BABEL-TEACH dataset is reported. 
}
\centering
\begin{tabular}{rcccccl}
\toprule
\multirow{2}{*}{Method} & \multicolumn{2}{c}{Pre-process} & \multirow{2}{*}{L2-P $\downarrow$} & \multirow{2}{*}{L2-Q $\downarrow$} & \multirow{2}{*}{NPSS $\downarrow$} \\
\cmidrule(l){2-3}
& Rand. Proj. & Scale & &  \\
\midrule
Interp. & - & - & 0.0764 & 0.0497 & 0.5501 \\
MDM~\cite{tevet2023mdm} & \xmark & - & 0.2236 & 0.1422 & 1.1508 \\
MDM~\cite{tevet2023mdm} & $\checkmark$ & - & 0.1952 & 0.1264 & 1.1509 \\
GMD~\cite{karunratanakul2023gmd} & $\checkmark$ & 0. & 0.3615 & 0.1410 & 1.4928 \\
GMD~\cite{karunratanakul2023gmd} & $\checkmark$ & 0.2 & 0.2917 & 0.1240 & 1.2498 \\
GMD~\cite{karunratanakul2023gmd} & $\checkmark$ & 0.5 & 0.2640 & 0.1153 & 1.1171 \\
GMD~\cite{karunratanakul2023gmd} & $\checkmark$ & 1.0 & 0.2502 & 0.1107 & 1.0816 \\
GMD~\cite{karunratanakul2023gmd} & $\checkmark$ & 2.0 & 0.2663 & 0.1182 & 1.2482 \\
\midrule
MMDM (Ours) & \xmark & - & \underline{0.0606} & \textbf{0.0359} & \textbf{0.2764} \\
MMDM (Ours) & $\checkmark$ & - & \textbf{0.0581} & \underline{0.0377} & \underline{0.2788} \\
\bottomrule
\end{tabular}
\label{tab:mib_ab}
\end{table}

\subsection{Additional qualitative comparisons}

We present additional qualitative comparisons using video clips, where we evaluate our model against three state-of-the-art methods: the motion in-betweening model CMIB~\cite{kim2022CMIB} and the motion generation models MDM~\cite{tevet2023mdm} and GMD~\cite{karunratanakul2023gmd}. Each competitor is optimized to deliver its best possible results. We provide 16 visual examples from the above methods, combined with the ground truth and our model's results, for a side-by-side comparison. All visual examples are randomly sampled from the testing set. Inside the video, from left to right, top to bottom, it displays the ground truth, CMIB~\cite{kim2022CMIB}'s result, MDM~\cite{tevet2023mdm}'s result, GMD~\cite{karunratanakul2023gmd}'s result and our MMDM's result. An indicator marks the preceding part, the transitioning part, and the succeeding part. These video samples can be found in the supplementary video. 

\end{document}